%% file: main.tex
\documentclass{article}
\usepackage{amssymb}
\usepackage{amsmath}
\usepackage{booktabs}
\usepackage{makecell}
\usepackage{wrapfig}
\usepackage[table]{xcolor}

\usepackage[final]{corl_2025} 

\usepackage{xcolor}
\definecolor{brickred}{HTML}{D94F30}
\definecolor{goldyellow}{HTML}{F2C14E}
\definecolor{olivegreen}{HTML}{90BE6D}
\definecolor{softblue}{HTML}{577590}
\usepackage{graphicx}

\title{Robot Learning from Any Images}

\author{
Siheng Zhao$^{1}$\thanks{Equal contribution. Author order determined by coin flip.} ~, Jiageng Mao$^{1*}$, Wei Chow$^{1}$, Zeyu Shangguan$^{1}$, Tianheng Shi$^{1}$,\\ \textbf{Rong Xue$^{1}$, Yuxi Zheng$^{1}$, Yijia Weng$^{2}$, Yang You$^{2}$,}\\ \textbf{Daniel Seita$^{1}$, Leonidas Guibas$^{2}$, Sergey Zakharov$^{3}$, Vitor Guizilini$^{3}$, Yue Wang$^{1}$}\\
$^{1}$University of Southern California, $^{2}$Stanford University, $^{3}$Toyota Research Institute
}

\begin{document}
\maketitle
\vspace{-6mm}
\begin{abstract}
    We introduce RoLA, a framework that transforms any in‑the‑wild image into an interactive, physics‑enabled robotic environment. 
    Unlike previous methods, RoLA operates directly on a single image without requiring additional hardware or digital assets.
    Our framework democratizes robotic data generation by producing massive visuomotor robotic demonstrations within minutes from a wide range of image sources, including camera captures, robotic datasets, and Internet images.    
    At its core, our approach combines a novel method for single-view physical scene recovery with an efficient visual blending strategy for photorealistic data collection. We demonstrate RoLA's versatility across applications like scalable robotic data generation and augmentation, robot learning from Internet images, and single-image real-to-sim-to-real systems for manipulators and humanoids. Video results are available at our \href{https://sihengz02.github.io/RoLA/}{project page}.
 
\end{abstract}


\input{sections/intro}
\input{sections/related}

\input{sections/method}
\input{sections/exp}

\input{sections/conclusion}

\clearpage
\input{sections/limitation}



\bibliography{main}  

\clearpage
\input{sections/appendix}

\end{document}

%% file: sections/intro.tex
\vspace{-4mm}
\begin{center}
    \textit{``Each photograph is a portal to the physical world.''}
\end{center}

\begin{figure}[!ht]
\vspace{-3mm}
\centering
\includegraphics[width=0.99\linewidth]{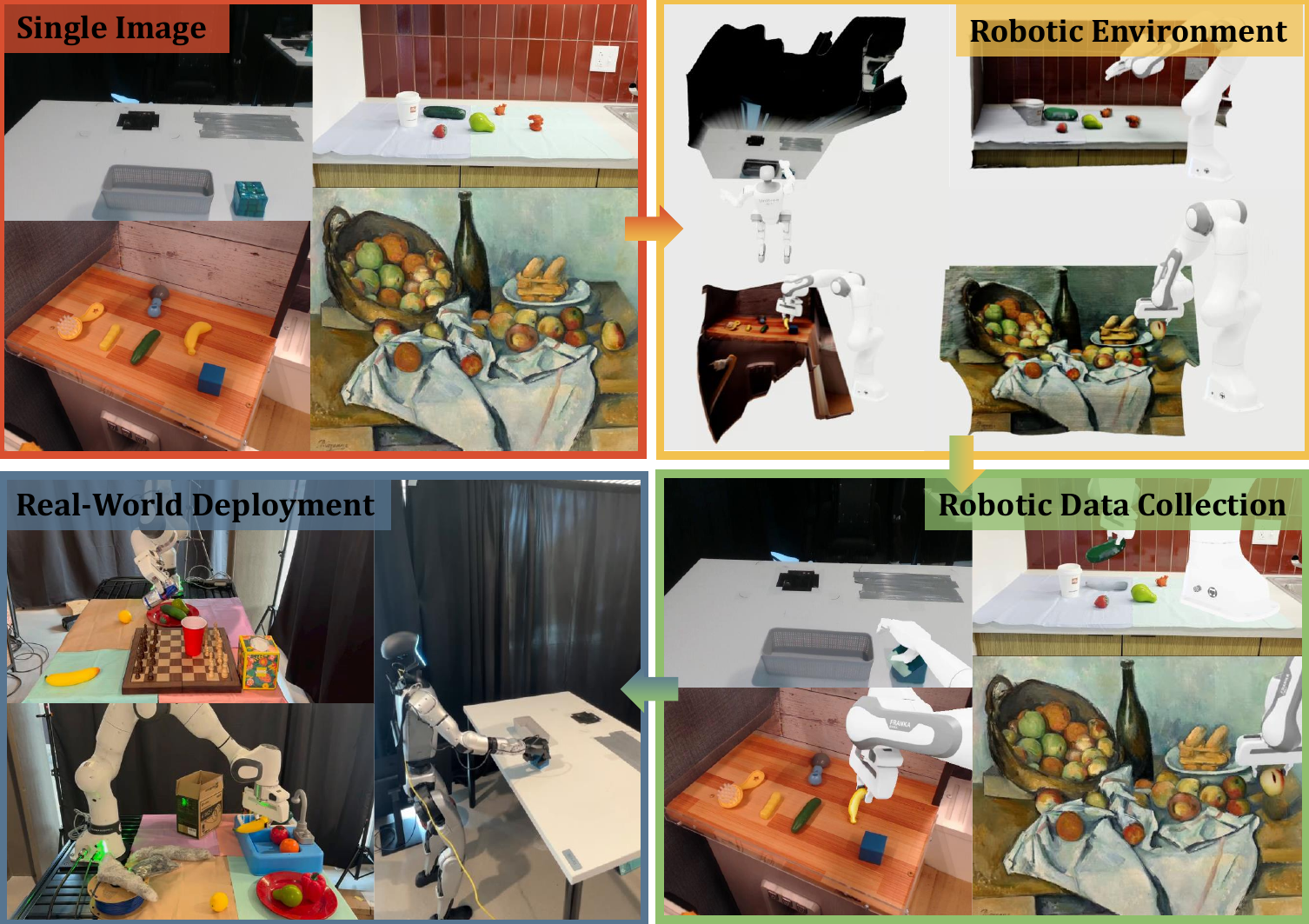}
\vspace{-1mm}
\caption{RoLA transforms a single in-the-wild image into an interactive, physics-enabled robotic environment.
Given a single input image (\textbf{\textcolor{brickred}{top-left}}), RoLA recovers the physical scene for robot learning (\textbf{\textcolor{goldyellow}{top-right}}), enables large-scale robotic data generation (\textbf{\textcolor{olivegreen}{bottom-right}}), and supports deployment of learned policies on real robots (\textbf{\textcolor{softblue}{bottom-left}}).}
\label{fig/teaser}
\vspace{-3mm}
\end{figure}

\section{Introduction}

Data quantity and diversity are widely recognized as primary bottlenecks in scaling robot learning. Although on‑robot demonstrations provide physically accurate supervision, collecting them at scale demands specialized hardware and extensive human effort. Conversely, non‑robotic visual data, such as images and videos, is virtually unlimited and encodes rich information relevant to robotic tasks. However, converting raw pixels into robot-complete data, \textit{i.e.}, structured robotic scenes and demonstrations that can be directly used for policy learning, remains a fundamental challenge.

Real‑to‑sim‑to‑real methods~\citep{acdc,grs,re3sim,real2sim,rebot, robogs,robogsim,scalable-real2sim,video2policy,vlm-real2sim,vr-robo} offer a promising direction toward bridging this gap. These approaches reconstruct robotic environments from multiview images or videos, simulate interactions to collect training data, and subsequently transfer the learned policies to real-world robots. While effective, their reliance on \textit{ad hoc} hardware setups confines their data collection to laboratory settings, limiting their scalability to massive in-the-wild images that are readily available on the Internet or in the real world.  

This paper poses a fundamental question: \textit{Can we obtain robot-complete data from non-robotic images under minimal assumptions—ideally from a single image?} To answer this question, we revisit the underlying reasons why prior works rely on \textit{ad hoc} camera setups to build robotic environments: (1) to reconstruct the scene and object geometry as simulation assets; (2) to generate photorealistic visual observations through physics-based or image-based rendering. Our key insight is that multiview reconstruction may not be necessary for building robotic scenes, as with foundation model priors, a single RGB image is sufficient for recovering a physical scene. Furthermore, high-quality visual observations can be generated without full rendering pipelines, as visual blending of real images and virtual assets can also achieve high realism. Building on these insights, we eliminate the special need for hardware setups and make real-to-sim-to-real pipelines generalizable to arbitrary single images from anywhere without requiring additional information.

To this end, we introduce RoLA (Robot Learning from Any images), which automatically recovers a physical robotic scene configuration from a single image and enables scalable, photorealistic visuomotor data collection for policy learning. The framework consists of three key components: (1) Real-to-Sim: Recovers the physical scene from a single image by estimating object and scene geometry, inferring physical properties, establishing scene-object relationships, recovering camera pose, and placing the robot in appropriate locations; (2) Simulation: Generates large-scale simulated robotic trajectories in the recovered physical scene for diverse robotic tasks; (3) Sim-to-Real: Synthesizes photorealistic visuomotor demonstrations via visual blending for policy learning and deploys the learned policies to real robots. Together, these components form a fully automatic and generalizable pipeline that transforms any single image into a physically grounded, robot-interactable scene, bridging the gap between passive visual data and embodied robotic action.

By unleashing the power of in-the-wild images, RoLA unlocks a wide range of applications in real-world robotics. (1) Given a single image from a robotic camera, RoLA can recover the underlying physical scene and generate unlimited, physically plausible visuomotor demonstrations, providing abundant data for training imitation learning policies. (2) When paired with automatically generated language descriptions, RoLA further facilitates scalable data generation and augmentation for training vision-language-action (VLA) models. 
(3) RoLA also enables leveraging broader in-the-wild images, such as Internet photos, as priors for learning vision-based robotic manipulation. Using apple picking as a case study, we show that manipulation performance improves by incorporating priors learned from large-scale non-robotic visual data.
 

%% file: sections/related.tex
\section{Related Works}

\textbf{Real-to-Sim-to-Real.} Creating digital twins from real-world data offers a promising pathway to scale robot learning in simulation. Prior efforts have explored building digital robotic environments through multiview scene reconstruction~\citep{real2sim, scalable-real2sim, vlm-real2sim, re3sim, vr-robo, matterport3d, gibson, robogs,robogsim,tiebot}, object reconstruction~\citep{urdformer,ditto,structurefromaction,real2code}, procedural generation~\citep{behavior1k, ai2thor, holodeck, robogen, robocasa}, or assets retrieval~\citep{acdc}. However, most works need manual scene setup or require multiview camera captures, making them less generalizable to diverse real-world data. Notably, some works~\citep{acdc, grs} focus on recovering the 3D scene from a single image. Nonetheless, these works typically retrieve existing assets, thus constraining generalization to the scope of pre-existing databases. In contrast, our approach is fully automatic and operates on any single image without requiring additional information or external databases.

\textbf{Robot Learning from Unstructured Data.} Unstructured Internet images and videos offer a vast but underutilized resource for training robot policies. Prior work extracts actionable knowledge from videos by learning priors from human-object interactions~\citep{handobject, object-dex, human-to-robot}, motion retargeting~\citep{uh1, omnih2o}, inverse dynamics~\citep{unisim, actionless-video}, and affordance transfer~\citep{ram, vidbot}. Notably, recent works resort to pixel-level augmentations, e.g., changing backgrounds~\citep{roboengine}, viewpoints~\citep{rovi}, embodiments~\citep{phantom, shadow, ar2d2}, and green screens~\citep{simplerenv, rebot, greenscreen}, to synthesize diverse visual demonstrations. However, these pixel editing approaches cannot ensure physical accuracy in generating robotic demonstrations. In contrast, our approach combines physics simulation and visual blending, which generates both physically accurate and photorealistic robotic demonstrations.

\textbf{Physics-based Scene Generation.} Several recent works~\citep{physgen3d, phystwin, physcene, physbench} explore generating a physically interactable digital twin from a single image, typically for physics-based video generation~\citep{physgen, autovfx}. However, these works generally focus on narrow object categories and show limited applications in robotics. CAST~\cite{cast} is the most relevant work that also explores transforming a single image into an interactable digital twin. However, CAST doesn't model the full background and demonstrates limited applications in robotics. In contrast, our approach is generalizable to diverse images and objects, and fully supports robotic simulation and data generation.

%% file: sections/method.tex
\section{Method}
\begin{figure}[!t]
\vspace{-6pt}
\centering
\includegraphics[width=0.99\linewidth]{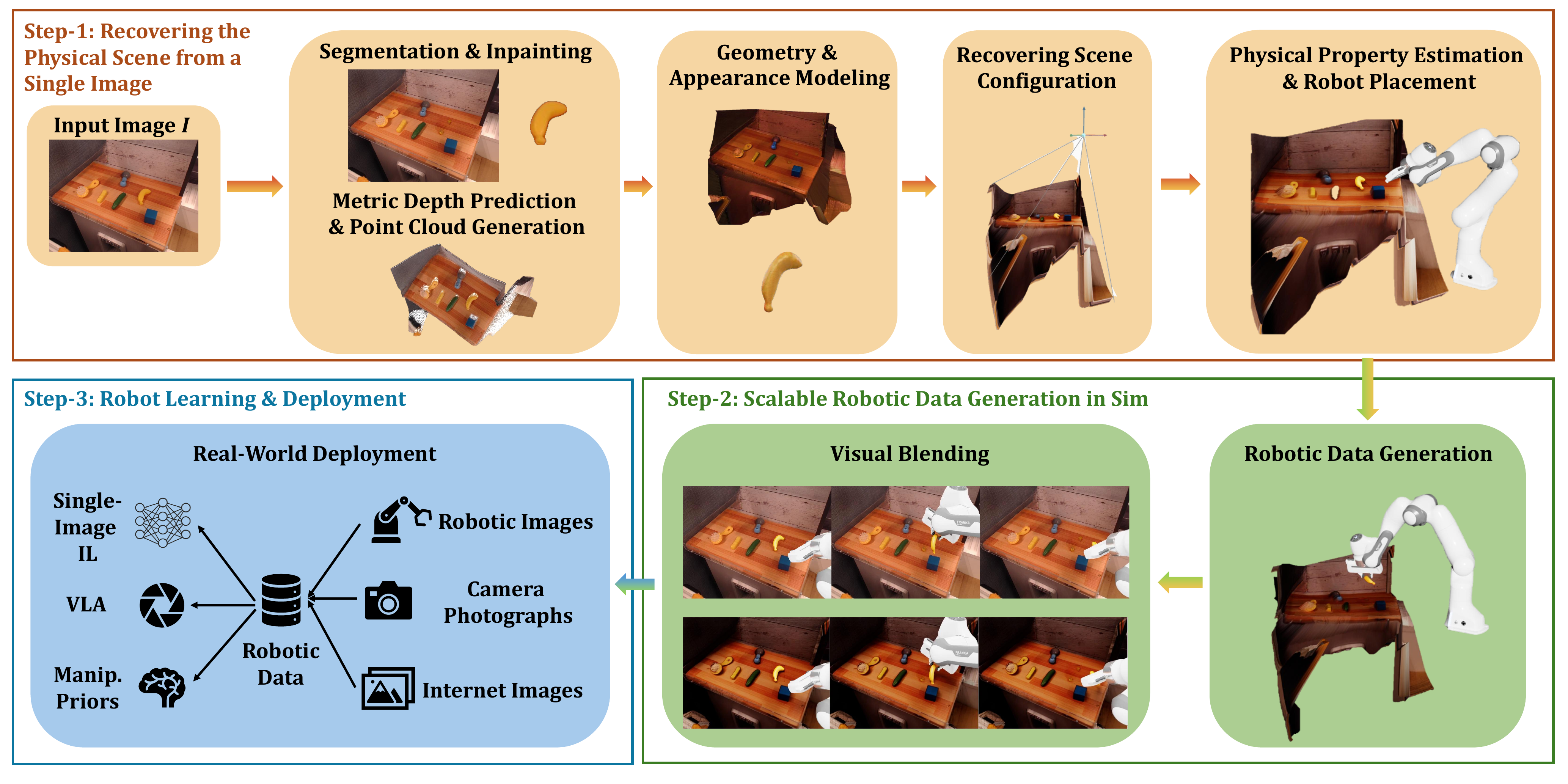}
\caption{An overview of the RoLA framework. \textbf{\textcolor[HTML]{AA4B18}{Step 1}}: Recover the physical scene from a single image. \textbf{\textcolor[HTML]{3B7D23}{Step 2}}: Generate large-scale photorealistic robotic demonstrations via visual blending. \textbf{\textcolor[HTML]{0B76A0}{Step 3}}: Train and deploy policies across tasks and embodiments using the collected data.}
\label{fig:method}
\vspace{-6pt}
\end{figure}

We study how to leverage a single image from any source as a robotic environment for scalable robotic data generation and policy learning. We achieve this goal by first recovering the physical scene behind the input image (Section~\ref{sec-3.1}), then collecting visuomotor robotic demonstrations in the physical scene (Section~\ref{sec-3.2}), and finally training robotic policies and deploying them to the real robots (Section~\ref{sec-3.3}). An overview of our method is shown in the Figure~\ref{fig:method}.

\subsection{Recovering the Physical Scene from a Single Image} \label{sec-3.1}

\textbf{Problem Definition.} We consider the inverse problem of how a real image is created. The forward image formation process can be expressed as:
\begin{equation}
    I = \pi(S; \mathcal{C}),
\end{equation}
where $I \in \mathbb{R}^{H \times W \times 3}$ is the observed image, and $\pi$ is the camera projection function. The physical scene $S = \{O, B\}$ consists of objects $O$ and a background $B$, each represented as a textured mesh $\mathcal{M}$ with central position $\mathbf{p}$, orientation $\mathbf{q}$ (e.g., quaternion), and physical properties $\mathcal{P}$ (such as mass and friction). The camera parameters $\mathcal{C}$ include intrinsics $\mathbf{K}$ and extrinsics $[\mathbf{R} \mid \mathbf{t}]$.

Our goal is to recover the entire physical scene $\{\mathcal{M}, \mathcal{P}, \mathbf{p}, \mathbf{q}, \mathcal{C}\}$ from a single image $I$. This inverse problem is inherently ill-posed. However, we demonstrate that with generative priors, it is possible to obtain accurate and robust estimates of the underlying scene structure.

\textbf{Object and Background Modeling.} We begin by obtaining textured meshes $\mathcal{M}$ for both the objects $O$ and the background $B$, which recovers the scene geometry and appearance.

For object modeling, we first generate object segmentation masks from $I$ using the grounded segmentation model~\citep{groundedsam}. The segmented object regions are then passed to a state-of-the-art image-to-3D generation model~\citep{wonder3d}, which produces the object textured mesh $\mathcal{M}_{O}$. 

For background modeling, we use an image inpainting model~\citep{lama} to synthesize a background-only image $I_B$, which is used to texture the background mesh. To estimate background geometry, we apply a metric depth prediction model~\citep{depthpro} to infer an absolute depth map $D$ and camera intrinsics $\mathbf{K}$ from $I$. Using inverse projection, we construct a scene point cloud $P_S \in \mathbb{R}^{HW \times 3}$:
\begin{equation}
    \mathbf{X}_{uv} = D(u,v) \cdot \mathbf{K}^{-1} [u, v, 1]^\top,
\end{equation}
where $(u,v)$ are image coordinates and $\mathbf{X}_{uv}$ is the corresponding 3D point in the camera frame.

We then partition $P_S$ into object and background point clouds, $P_O$ and $P_B$, using the object segmentation masks. A background mesh $\mathcal{M}_B$ is reconstructed from $P_B$ via heightmap reconstruction. Since the background primarily serves as a support surface in robotic tasks, we can also approximate it using shape primitives (e.g., planes) to accelerate physics simulation.


\textbf{Recovering Scene Configuration.}  
Given the textured meshes $\mathcal{M} = \{\mathcal{M}_O, \mathcal{M}_B\}$ for the object and background, our goal is to recover the full scene configuration: the sizes, positions $\mathbf{p}$, and orientations $\mathbf{q}$ of the meshes, as well as the camera pose $[\mathbf{R} \mid \mathbf{t}]$ in a gravity-aligned world frame.

We leverage the scene point cloud $P_S$, derived from metric depth, as the reference geometry, since it inherently encodes both the scene’s spatial structure and its absolute scale. The background mesh $\mathcal{M}_B$, reconstructed directly from the background point cloud $P_B \subset P_S$, preserves its real-world scale and alignment. For the object mesh $\mathcal{M}_O$, which is generated independently from image-to-3D models, we rescale it and register its position $\mathbf{p}$ and orientation $\mathbf{q}$ to the corresponding object point cloud $P_O$ using mesh-to-point registration~\citep{icp}, which establishes the scene-object relationships.

To place the scene and camera in a physically meaningful world frame, we introduce a concept named the \textit{supported plane}, which is either the tabletop or the ground plane. We assume the \textit{supported plane} is always perpendicular to the Z-axis, i.e., the gravity axis in physics simulation. With this assumption, we first run a ground segmentation~\cite{groundedsam} and a RANSAC plane estimation algorithm on the scene point cloud $P_{S}$ to estimate the ground normal direction $\mathbf{n}$. We then estimate the rotations $\mathbf{R}$ of the scene and camera that maps the plane normal to the Z-axis $\mathbf{z}=[0,0,1]^{\!\top}$:
\begin{equation}
    \mathbf{R}
    \;=\;
    \mathbf{I}_3
    + \sin\theta\,[\mathbf{k}]_\times
    + (1-\cos\theta)\,[\mathbf{k}]_\times^2,    
\end{equation}
where $[\mathbf{k}]_\times$ is the $3\times3$ skew‑symmetric matrix of $\mathbf{k}=\frac{\mathbf{n}\times\mathbf{z}}{\|\mathbf{n}\times\mathbf{z}\|_2}$, and $\theta = \arccos(\mathbf{n}^{\!\top}\mathbf{z})$ following the Rodrigues’ formula. Finally, we rotate the scene and update the camera position $\mathbf{t}$ to center the scene at the origin of the world frame. This results in the full camera pose $[\mathbf{R} \mid \mathbf{t}]$ in the world frame.

This step fully recovers the spatial configuration of object, background, and camera, enabling physically plausible visuomotor demonstration generation from the original image layout and viewpoint.

\textbf{Physical Property Estimation.}
To enable physical interaction, we estimate key physical properties $\mathcal{P}$, such as mass (or density) and friction coefficients, required by physics simulators. 
To get initial estimates for different objects in a scalable way, we leverage large language models (LLMs)~\citep{gpt4o} to provide physical property estimates based on the object category and image context. Specifically, we first identify the object class, then prompt the LLM with the class name and visual context to infer plausible physical parameters.

\textbf{Robot Placement.}  
To convert the physical scene into a robotic environment, we must determine appropriate robot positions. For images captured by a robot-mounted camera, we directly use the known camera-to-robot transformation to place the robot in the correct pose. For non-robotic images, we propose a sampling-based method to generate feasible placements. We approximate the robot's reachable workspace as a 3D shell and sample base positions such that this region covers the axis-aligned bounding boxes of all objects. To avoid collisions, we discard placements that intersect with the scene’s bounding box. This process produces multiple valid placement candidates, enabling diverse action trajectory generation from different robot positions.

\subsection{Robotic Data Generation} \label{sec-3.2}

\textbf{Data Collection.}  
We support multiple modes of collecting robotic demonstrations within our image-generated environment. These include direct control of robots in physics simulators via keyboard or space mouse, scripted policies using motion planners~\cite{curobo}, and pretrained manipulation policies~\cite{anygrasp, graspnet}. These methods enable efficient large-scale robotic data collection; for example, over 1,500 visuomotor demonstrations can be collected within an hour using an RTX4090 GPU. For each demonstration, we record $T$ frames of robot actions $\{a_t\}_{t=0}^T$, depth maps $\{D_t\}_{t=0}^T$, and visual observations $\{I_t\}_{t=0}^T$ from the original camera viewpoint, which are later used to compose photorealistic visuomotor demonstrations.

\textbf{Visual Blending.}  
A key challenge in sim-to-real adaptation is the visual domain gap: rendered observations from simulators often differ significantly in appearance from real-world images. To mitigate this, we introduce a visual blending technique that enhances the photorealism of rendered visual observations by combining them with the real background image $I_B$ using z-buffering. 

Let $I_{B}$ and $D_{B}$ be the background image and depth, and $\{I_t\}_{t=0}^T$ and $\{D_t\}_{t=0}^T$ denote the rendered images and depth maps in simulation. The blended observations $\{I^{\prime}_t\}_{t=0}^T$ can be computed as
\begin{equation}
    I^{\prime}_t = M_t \odot I_t + (1 - M_t) \odot I_B,\quad\text{where}\quad M_t(u,v) = 
    \begin{cases}
        1, & \text{if } D_t(u,v) < D_B(u,v) - \epsilon, \\
        0, & \text{if } D_t(u,v) \geq D_B(u,v) - \epsilon.
    \end{cases}
\end{equation}
The binary mask $M_t \in \{0,1\}^{H \times W}$ indicates where the rendered surface is closer to the camera than the real background, $\epsilon$ is a small threshold for depth noise and alignment error, and $\odot$ denotes element-wise multiplication. This blending preserves the real image in background regions while inserting rendered content such as objects and robots only where it appears physically in front of the scene. Compared to~\citep{simplerenv}, visual blending with z-buffering handles the occlusion cases more effectively, resulting in more photorealistic and consistent visual observations. We store the resulting blended observations along with the corresponding robot actions $\{I^{\prime}_t, a_t\}^T_{t=0}$ as visuomotor demonstrations for downstream policy training.

\subsection{Robot Learning \& Deployment} \label{sec-3.3}

The image-generated robotic environments enable scalable data generation across a wide range of tasks and robotic embodiments. Using the collected visuomotor demonstrations $\{I^{\prime}_t, a_t\}^T_{t=0}$, we train visuomotor policies in simulation and seamlessly deploy them to the real world. Our framework can also be used as a scalable data generator to augment existing robotic datasets~\cite{bridgedata} by conditioning on their images as inputs. Furthermore, by automatically generating language annotations for visual demonstrations using LLMs, our method supports scalable robotic data generation for training more generalizable vision-language-action models (VLAs).

%% file: sections/exp.tex
\section{Experimental Evaluation}

\begin{figure}[!t]
\vspace{-6pt}
\centering
\includegraphics[width=0.99\linewidth]{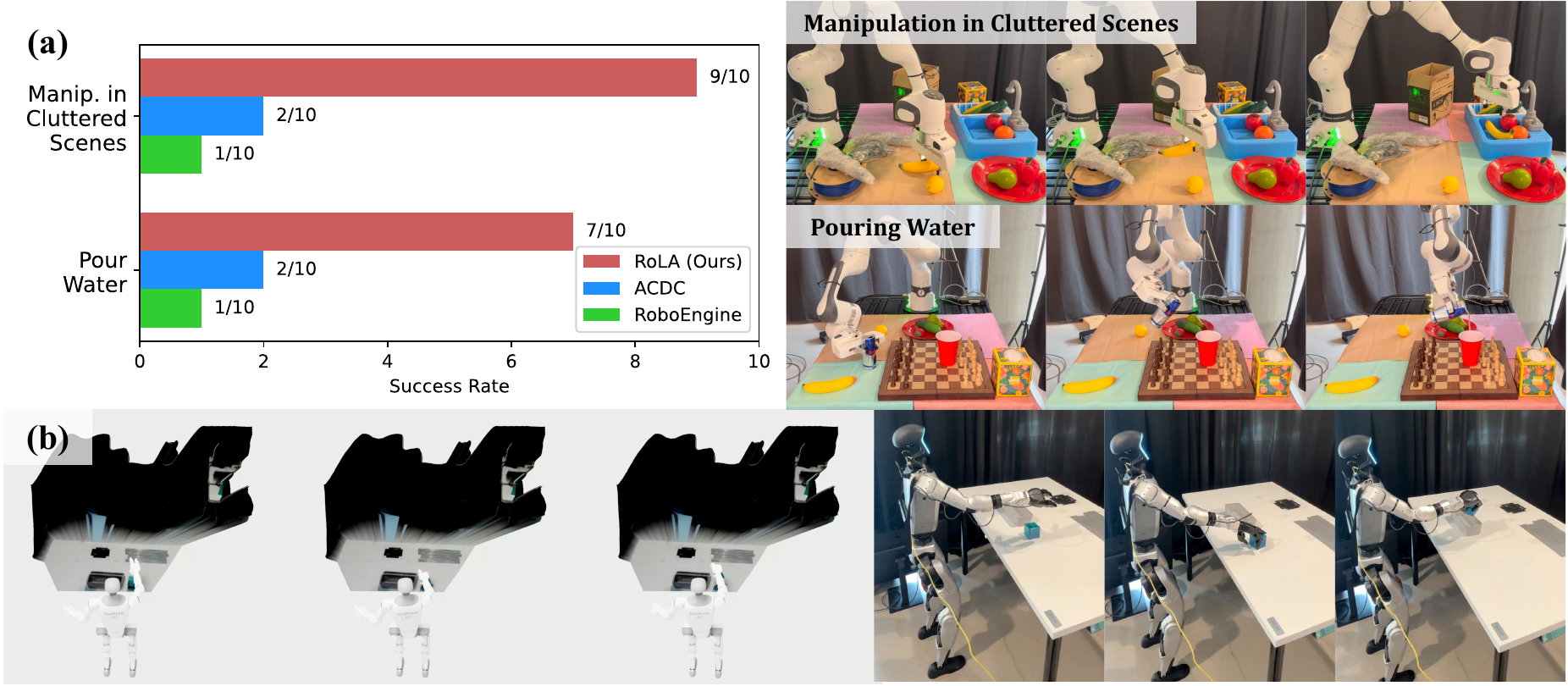}
\vspace{-2mm}
\caption{(a) Real-world deployment of policies trained with RoLA-generated data. (b) RoLA enables efficient real-to-sim-to-real transfer for humanoid robots.}
\label{fig:il_real_exp}
\vspace{-8pt}
\end{figure}

In this section, we investigate the following questions through extensive experiments:

\textbf{Q1:} Can RoLA recover an accurate physical scene from a single image that enables effective robotic policy learning, achieving performance comparable to  multiview reconstruction methods?
\\[0\baselineskip]
\textbf{Q2:} Can RoLA efficiently generate large-scale, high-quality robotic data? How does the quality of the generated data compare to previous retrieval-based and pixel-editing-based methods?
\\[0\baselineskip]
\textbf{Q3:} Can RoLA enable training of visuomotor policies and vision-language-action (VLA) models from single images, generalize across different embodiments (\textit{e.g.}, manipulators and humanoid robots), and deploy successfully to the real world?
\\[0\baselineskip]
\textbf{Q4:} Can RoLA leverage in-the-wild Internet images for vision-based robot learning?
\\[0\baselineskip]
\textbf{Q5:} How does visual blending improve RoLA’s performance?

Due to the page limit, we refer readers to the appendix for detailed experimental setups.

\subsection{Physical Scene Recovery}

\begin{wraptable}{r}{0.45\textwidth}
\vspace{-12pt}
\centering
\scriptsize
\begin{tabular}{l|c|c}
\toprule
\textbf{Method} & \textbf{Multi-view} & \textbf{Single-view (RoLA)} \\
\midrule
\textbf{Success Rate} & $75.5 \pm 7.5$\% & $72.2 \pm 2.1$\% \\
\bottomrule
\end{tabular}
\caption{Comparison of policy success rates between multi-view reconstruction and our single-view RoLA pipeline. Average success rates $\pm$ StdErr are calculated over 3 seeds.}
\label{tab:multiview_ablation}
\end{wraptable}

A common concern of single-image scene recovery is the potential inaccuracy of scene properties due to the lack of geometric constraints. We investigate this issue by comparing our method with a multiview reconstruction baseline~\cite{real2sim}. Specifically, we build digital twins of robotic environments using both the RoLA pipeline and~\cite{real2sim} on the task \textit{``pick up the banana and put it onto the stove''}, collect 200 demonstrations respectively, train an imitation learning policy~\citep{diffusionppolicy}, and deploy it for robotic manipulation. As shown in Table~\ref{tab:multiview_ablation}, RoLA's success rate closely matches the multiview reconstruction baseline, indicating that a single image is sufficient for creating a realistic robotic environment for policy learning (\textbf{Q1}). Although multiview reconstruction methods achieve slightly better performance ($\approx 3.3$\%), their reliance on controlled setups makes them significantly less scalable than our approach, which operates on single images without additional requirements.

\subsection{Robotic Data Generation} \label{sec:3.2}

To evaluate RoLA’s data generation quality, we set up a robotic manipulation benchmark and compare policies trained on RoLA-generated data with those from baseline methods: (1) ACDC~\citep{acdc}, a retrieval-based single-image data generation method, and (2) RoboEngine~\citep{roboengine}, an image-based demonstration augmentation approach. For each task, we collect 200 visuomotor demonstrations using RoLA and each baseline, train imitation learning policies~\citep{diffusionppolicy}, and measure success rates. As shown in Table~\ref{tab:baseline}, RoLA significantly outperforms the baselines across tasks, indicating that data generation from recovered physical scenes yields more faithful and physically accurate demonstrations than retrieval or pixel-editing methods (\textbf{Q2}).

\input{tables/il_baseline}

\subsection{Single-Image Robot Learning} \label{sec:4.3}

We conduct real robot experiments to investigate whether RoLA can enable single-image real-world robot learning. Specifically, we employ a Franka Research 3 Robot as the tested embodiment. 
\input{tables/vla}

We set up 2 real-world robotic manipulation tasks across diverse environments: (1) Object manipulation in cluttered scenes: the robot must pick up a banana in a cluttered environment and place it accurately inside a sink, evaluating grasp accuracy and collision avoidance. (2) Pouring water: The robot must pour soda from a can into a cup, evaluating manipulation precision. For each task, we capture an RGB image of the scene using a RealSense D415 camera as input to RoLA, then collect 200 demonstrations using our framework, train a diffusion policy, and deploy it to the real robot. During evaluation, we test each task for 10 trials. In each trial, the manipulated object is randomly placed within a 6cm$\times$6cm workspace with random poses. As shown in Figure~\ref{fig:il_real_exp}(a), the policies trained with RoLA can be successfully deployed to the real world with high success rates.

To further investigate whether RoLA supports different embodiments, \textit{e.g.}, a humanoid robot, we leverage a Unitree G-1 robot as the tested embodiment. We leverage a photo captured by the humanoid's head-mounted camera as input to RoLA. 
Through the RoLA pipeline, we reconstruct the physical scene from the ego-centric image, control the robot to grasp a blue cube with a dexterous hand in the recovered scene, and deploy the actions to the real robot to complete the same task. 
As shown in Figure~\ref{fig:il_real_exp}(b), the humanoid robot successfully finished the task, demonstrating RoLA’s support for different robot types. Experiments on both robotic manipulators and humanoid robots have demonstrated that RoLA effectively supports single-image, real-world robot learning (\textbf{Q3}).

\subsection{Training Vision-Language-Action Models with RoLA-Generated Data}

\begin{wrapfigure}{r}{0.5\textwidth}
\vspace{0pt}
\centering
\includegraphics[width=1.\linewidth]{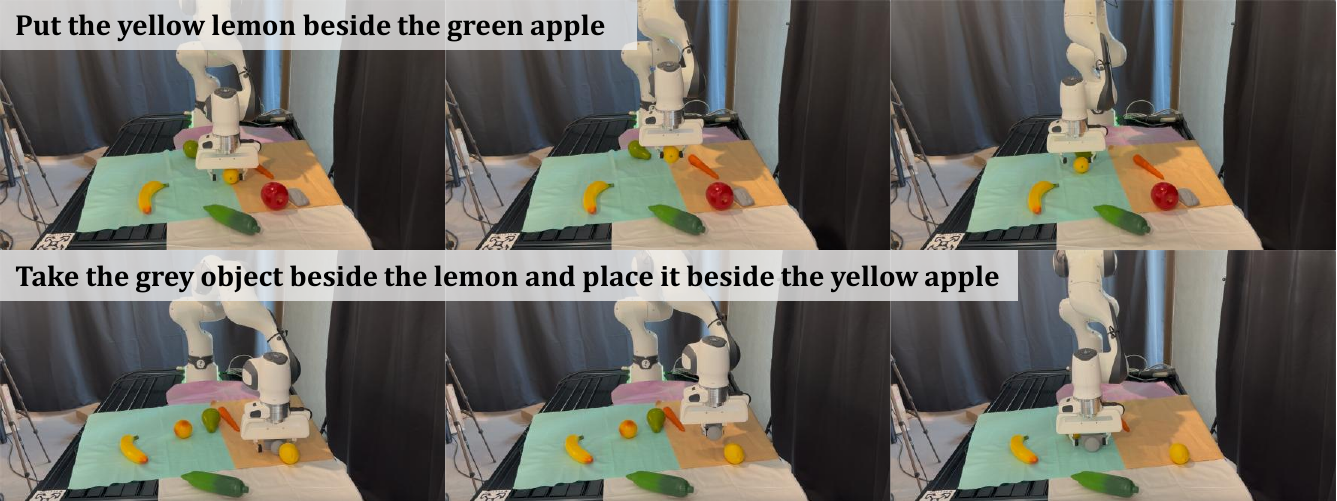}
\caption{Real-world execution of a VLA model trained with RoLA-generated data.}
\label{fig/vla_real}
\end{wrapfigure}

RoLA supports highly efficient and scalable robotic data generation from single images. We investigate whether this can facilitate large-scale training of vision-language-action (VLA) models. To this end, we collect over 60,000 demonstrations (the same scale of BridgeData V2~\citep{bridgedata}), and train a VLA model from scratch using Qwen2.5~\citep{qwen} VLM as the backbone, following the MiniVLA~\citep{minivla} training setting. Training occurs on 8$\times$H100 GPUs over four days (i.e. 768 GPU-hours).

\begin{figure}[!ht]
\vspace{-6pt}
\centering
\includegraphics[width=0.99\linewidth]{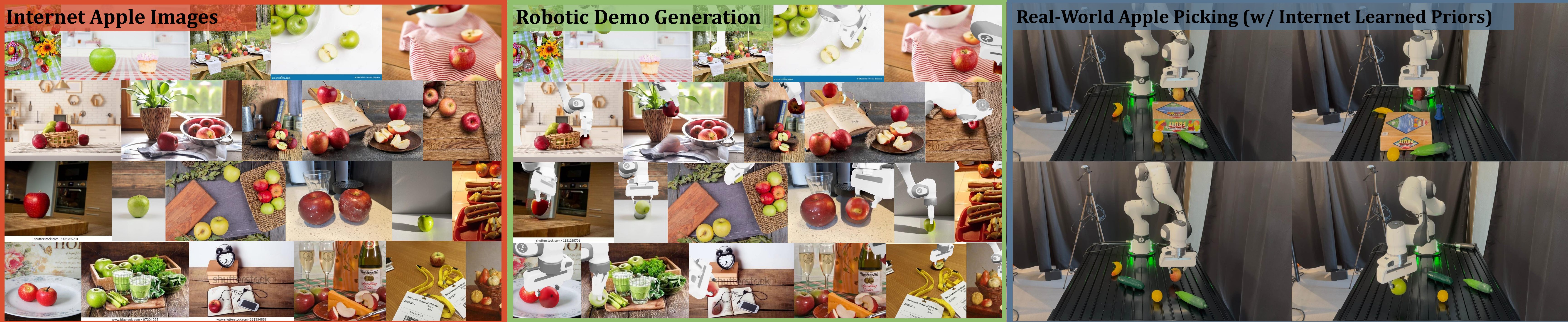}
\vspace{-2mm}
\caption{Learning a vision-based apple grasping prior from Internet apple images.}
\label{fig/apple}
\vspace{-14pt}
\end{figure}

To evaluate our VLA model’s performance, we design 10 evaluation tasks under the widely-used SimplerEnv~\citep{simplerenv} setting, testing skill diversity, scene variations, object category and pose variations, semantic understanding, and complex visual reasoning. Each task is evaluated over 10 trials with randomly initialized object poses. Table~\ref{tab:vla} illustrates our model's strong generalization from purely RoLA generated data. We then successfully deploy the VLA model to real-world scenarios, with example rollouts showcased in Figure~\ref{fig/vla_real}.

\subsection{Learning a Vision-Based Grasping Prior from Internet Images}

\begin{wraptable}{r}{0.4\textwidth}
\vspace{-6pt}
\centering
\scriptsize
\begin{tabular}{l|ccc}
\toprule
\textbf{Method} \textbackslash \textbf{\# of Demos} & \textbf{10} & \textbf{20} & \textbf{50} \\
\midrule
\textbf{w/ grasping prior} & 2/10 & 3/10 & 8/10 \\
\textbf{w/o grasping prior} & 0/10 & 3/10 & 3/10 \\
\bottomrule
\end{tabular}
\caption{Real-world apple grasping success across varying numbers of finetuning demonstrations, w/ and w/o the Internet-trained grasping prior.}
\label{tab:prior}
\end{wraptable}

We explore the potential of more in-the-wild data, such as Internet images, as a pretraining source for robot learning. Due to resource limitations, we only chose the apple grasping task as a case study. Specifically, we collect 2,000 Internet apple images and leverage RoLA to generate over 3,000 demonstrations to pick up apples in the Internet images (See Figure~\ref{fig/apple}). We pre-train a neural network that encodes images and outputs the 6-DoF grasping pose in the image 3D frame. Since Internet images contain apples of various sizes, types, positions, and with diverse backgrounds and viewpoints, we hope the pretrained network can learn a generalizable vision-based apple grasping prior from the Internet images. During deployment, we finetune the network with 10/20/50 real demonstrations and test the real-world apple grasping success rate with different apple types, positions, and camera viewpoints. As shown in Table~\ref{tab:prior}, the grasping success rate significantly increases with the learned grasping prior, indicating the potential of pretraining with Internet images for robot learning (\textbf{Q4}).

\subsection{Empirical Study}

\input{tables/ablation}

We investigate how the proposed visual blending affects the learning performance. In particular, we follow the training setting in Section~\ref{sec:3.2} but replace visual blending with rendering to generate visuomotor demonstrations. As shown in Table~\ref{tab:blending_ablation}, visual blending significantly improves the visual quality of demonstrations and enhances the learning performance (\textbf{Q5}).

%% file: tables/il_baseline.tex
\setlength{\tabcolsep}{5pt}
\begin{wraptable}{r}{0.6\textwidth}
\vspace{-12mm}
\centering
\scriptsize
\begin{tabular}{l|ccc}
\toprule
\textbf{Task} &  \textbf{RoLA (Ours)} & \textbf{RoboEngine} & \textbf{ACDC}\\
\midrule
\textbf{Broccoli into Bowl} & $53.4 \pm 11.7$\%  & $8.3 \pm 11.2$\%  &  $25.1 \pm 17.5$\% \\
\textbf{Banana onto Stove} & $85.4 \pm 11.6$\%  &  $3.3 \pm 4.7$\% & $0.0 \pm 0.0$\% \\
\textbf{Carrot onto Burner} &  $90.0 \pm 10.8$\% &  $22.7 \pm 16.6$\%  & $6.4 \pm 9.0$\% \\
\midrule
\rowcolor{gray!20}
\textbf{Average} & \textbf{\boldmath$76.2 \pm 19.8$\%}  &  $11.4 \pm 13.4$\% & $10.5 \pm 14.9$\% \\
\bottomrule
\end{tabular}
\caption{Comparison of imitation learning success rates using data generated by RoLA versus baselines. Average success rates $\pm$ StdErr are calculated over 3 seeds.}
\vspace{-2mm}
\label{tab:baseline}
\end{wraptable}

%% file: tables/vla.tex
\renewcommand{\arraystretch}{1.1}
\begin{table}[t!]
\vspace{-5mm}
\centering
\scriptsize
\begin{tabular}{lc|lc}
\toprule
\textbf{Language Instruction} & \textbf{Success Rate} & \textbf{Language Instruction} & \textbf{Success Rate} \\
\hline
reach the blue cup & 10/10 & pick up the carrot & 10/10 \\
\hline
put the green pepper beside the lemon & 9.5/10 & put the yellow lemon beside the green apple & 10/10 \\
\hline
\makecell[l]{take the banana and place it\\ beside the potato} & 8/10 & \makecell[l]{pick up the red tomato and place it\\ beside the yellow lemon} & 9/10 \\
\hline
\makecell[l]{take the grey object beside the lemon\\ and place it beside the yellow apple} & 6.5/10 & \makecell[l]{take the strawberry off the potato and\\ place it beside the red apple} & 10/10 \\
\bottomrule
\end{tabular}
\vspace{6pt}
\caption{Simulation evaluation of our VLA model trained on RoLA-generated data. Each task is tested over 10 trials with randomized object poses, under SimplerEnv~\citep{simplerenv} setting. Note that \textit{partial success} (score of 0.5) is possible for tasks following OpenVLA~\citep{openvla}. See Appendix for more details.}
\label{tab:vla}
\vspace{-12pt}
\end{table}

%% file: tables/ablation.tex
\setlength{\tabcolsep}{4pt}
\begin{wraptable}{r}{0.5\textwidth}
\vspace{-12mm}
\centering
{\fontsize{7pt}{7.5pt}\selectfont
\begin{tabular}{l|c|cc}
\toprule
\textbf{Task} & \textbf{Robot} & \textbf{w/ Blending} & \textbf{w/o Blending} \\
\midrule
Pickplace Banana & WidowX & $80.0 \pm 17.8$\% & $33.3 \pm 23.5$\% \\
Spatula onto Board & WidowX & $85.0 \pm 4.0$\% & $6.7 \pm 9.4$\% \\
Broccoli into Bowl & WidowX & $53.4 \pm 11.7$\% & $0.0 \pm 0.0$\% \\
Pickplace Spoon & Franka & $100.0 \pm 0.0$\% & $22.2 \pm 22.0$\% \\
Banana onto Stove & Franka & $85.4 \pm 11.6$\% & $11.7 \pm 16.5$\% \\
Carrot onto Burner & Franka & $90.0 \pm 10.8$\% & $0.0 \pm 0.0$\% \\
\bottomrule
\end{tabular}
\caption{Ablation study comparing visual blending with direct mesh rendering (w/o blending). Visual blending yields better performance across both WidowX and Franka robots.}
\label{tab:blending_ablation}
}
\end{wraptable}

%% file: sections/conclusion.tex
\section{Conclusion}

We presented RoLA, a framework that transforms a single image into an interactive, physics-enabled robotic environment for scalable data generation and policy learning. RoLA faithfully recovers physical scenes from diverse image sources, enabling the collection of photorealistic, physically plausible demonstrations without multiview captures. Through extensive robotic experiments, we demonstrate that RoLA supports single-image robot learning, real-world deployment, and training of vision-language-action models. Our results highlight the potential of leveraging vast in-the-wild visual data to advance scalable, generalizable robot learning. 

%% file: sections/limitation.tex
\section{Limitations}

This work has several limitations. (1) The quality of the generated robotic demonstrations is bounded by the fidelity of the physics simulator. However, we emphasize that the core technical contribution of this paper lies in a highly scalable and generalizable approach for transforming massive non-robotic visual data into physically plausible robotic data. While the transformed demonstrations may not match the precision of real-world collected data, our framework can generate data several orders of magnitude faster and at a much larger scale than real-world collection. By leveraging the virtually unlimited pool of Internet images, RoLA supports data generation with high diversity in object types, scene layouts, appearances, and viewpoints, significantly expanding the scalability and generalization potential of robot learning. (2) While our method does not support changing camera viewpoints within a single scene as multiview reconstruction methods do, it compensates by leveraging Internet images captured from diverse natural viewpoints. As a result, RoLA can generate training data with rich viewpoint diversity across scenes, enabling the learned policies to generalize well to novel camera perspectives.
(3) Since we rely on generative priors to recover physical scenes from partially observable scans, the generated shapes may not be well aligned with the observation and may cause collisions in physics simulation. 

\section{Acknowledgements}
The USC Geometry, Vision, and Learning Lab acknowledges generous supports from Toyota Research Institute, Dolby, Google DeepMind, Capital One, Nvidia, and Qualcomm. Yue Wang is also supported by a Powell Research Award.

%% file: sections/appendix.tex
\appendix


\section{Experiments}

\subsection{Physical Scene Recovery}

\begin{figure}[!h]
\centering
\includegraphics[width=0.99\linewidth]{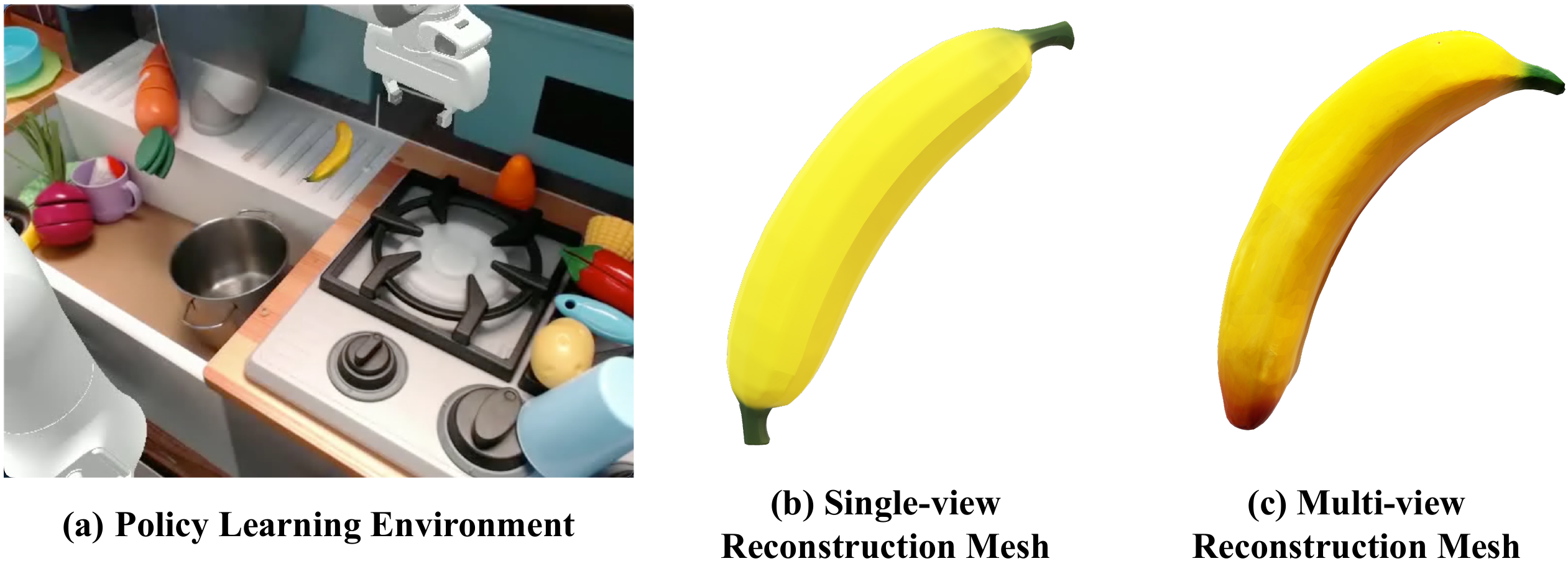}
\caption{Comparison of the policy learning environment and the reconstructed scene meshes from the single-view RoLA pipeline and multiview Polycam pipeline.}
\label{fig:appendix-view-comparison}
\end{figure}

The multiview method is conducted by capturing a short video of the scene using a handheld smartphone, then using Polycam\footnote{\url{https://poly.cam/}} to perform 3D reconstruction from the video. Polycam processes the frames to generate a textured mesh of the environment, which is then used to extract object geometry for simulation. Figure~\ref{fig:appendix-view-comparison} shows the policy learning environment of the task, along with the reconstructed meshes from both the single-view (RoLA) and multiview (Polycam-based) pipelines.

\subsection{Robotic Data Generation}
\begin{wrapfigure}{r}{0.5\textwidth}
\centering
\includegraphics[width=1.\linewidth]{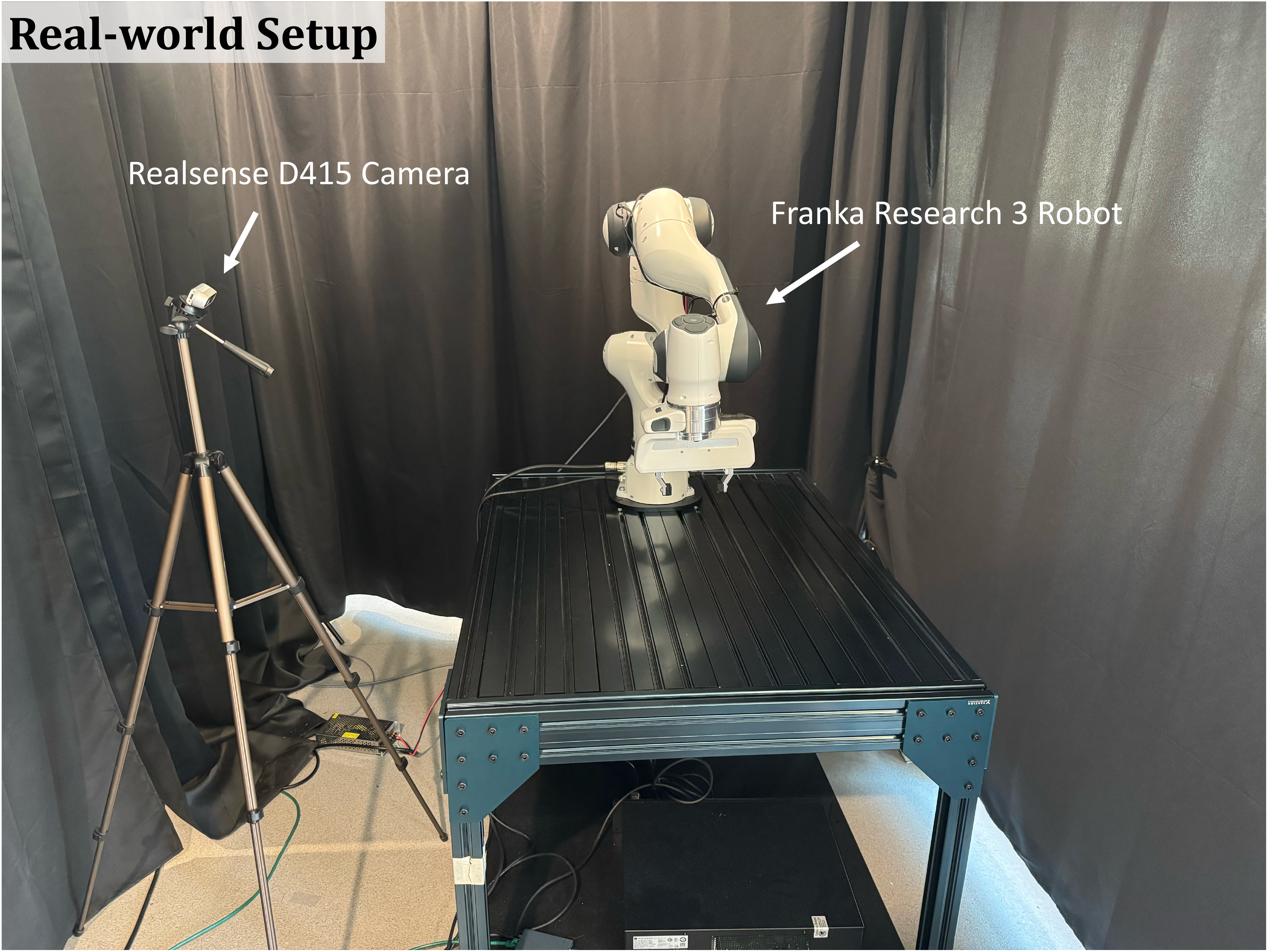}
\caption{Real-world experiment setup. We employ a Franka
Research 3 Robot and a RealSense D415 camera.}
\label{fig:appendix-setup}
\end{wrapfigure}

(1) ACDC\citep{acdc} is a retrieval-based, single-image data generation method. We follow its original pipeline with modifications to better suit our task. Given a single RGB image, ACDC performs four sequential steps. First, we extract per-object information by prompting GPT-4o\footnote{\url{https://chatgpt.com/}} to generate captions for all visible objects. These captions are then passed to GroundedSAM\citep{ren2024grounded} to produce object masks. Next, we compute the similarity between each masked object and entries in our object database using SigLip2~\citep{tschannen2025siglip}, retrieving the most visually similar ``digital cousin'' from OmniObject3D~\citep{wu2023omniobject3d}. In the third step, we use metric depth information to align each retrieved mesh with the corresponding object and post-process the matched assets to construct a fully interactive simulated scene. Finally, we employ the same demonstration collection pipeline as RoLA to generate 200 demonstrations per task for imitation policy learning, ensuring fair comparison.

(2) RoboEngine~\citep{roboengine} is an image-based demonstration augmentation method. We adopt the original codebase without modification. Given a single pre-collected demonstration, RoboEngine segments the robot arm and the manipulated object (i.e., the foreground), then uses a diffusion-based model to synthesize a physically plausible background conditioned on a foreground mask and a scene description. We apply this process to augment each demonstration into 200 demonstrations per task for imitation policy learning, enabling a consistent comparison with other methods.

\begin{figure}[!h]
\centering
\includegraphics[width=0.99\linewidth]{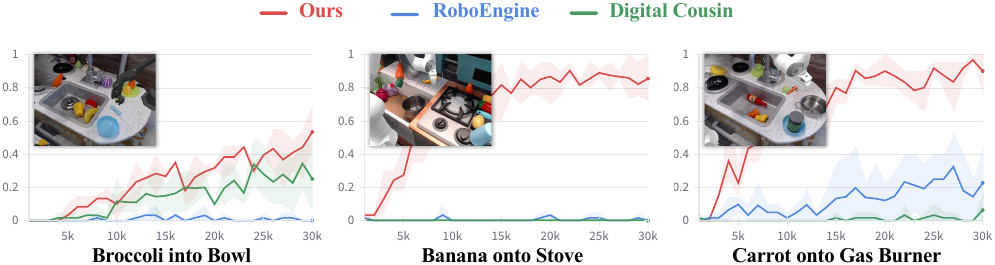}
\caption{Baseline comparison for robotic data generation.}
\label{fig:appendix-datagen}
\end{figure}

Figure~\ref{fig:appendix-datagen} shows the evaluation tasks (images on the left) and the comparison with the baseline methods. RoLA consistently outperforms the baselines in different tasks.

\subsection{Single-Image Robot Learning}

\begin{figure}[!h]
\centering
\includegraphics[width=0.99\linewidth]{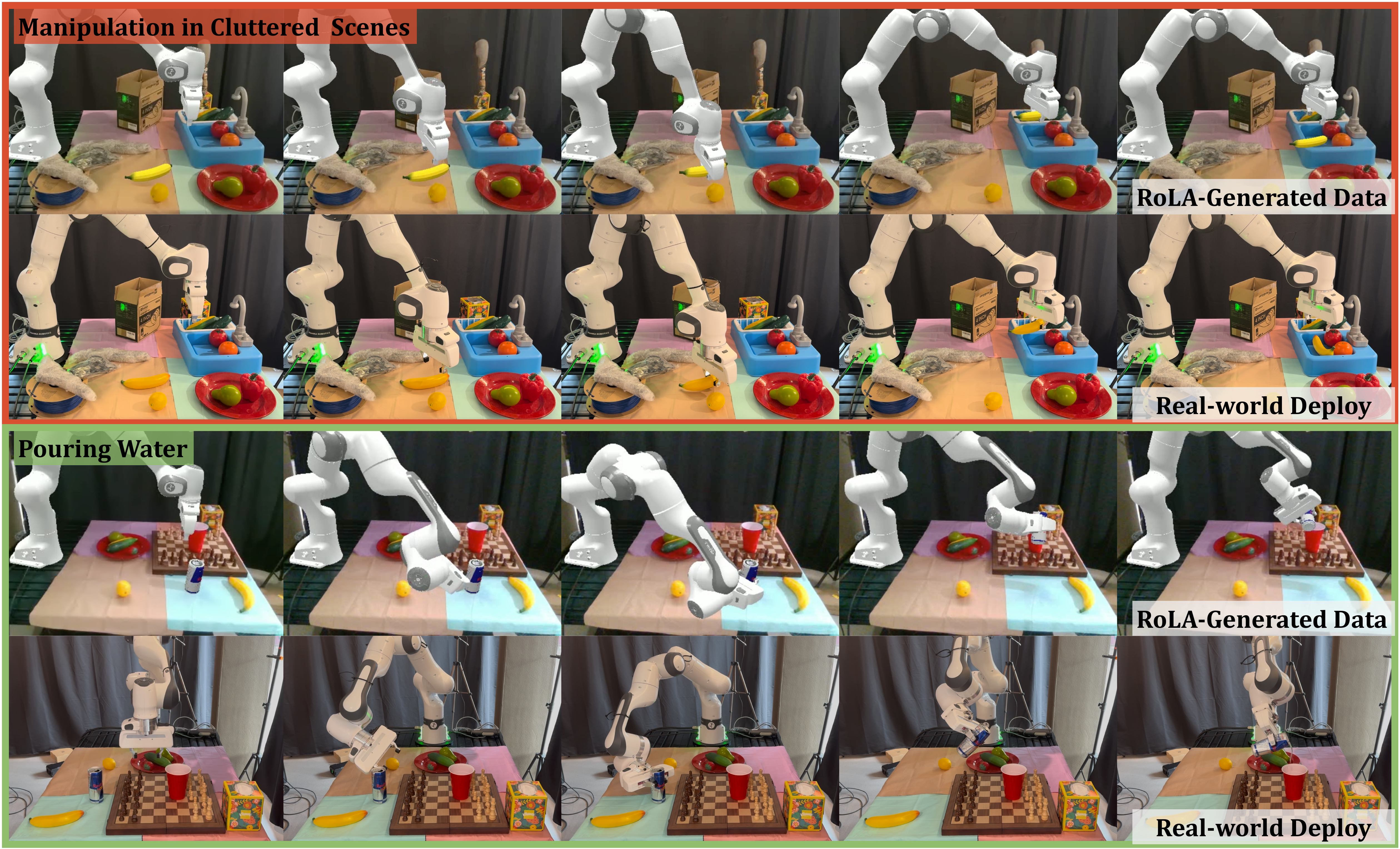}
\caption{\textbf{RoLA-Generated Data vs. Sim2Real Deployment.}
Each row pair shows the simulated RoLA-generated scenes (top) and corresponding real-world executions (bottom), demonstrating strong sim-to-real consistency in object placement, robot behavior, and overall task execution.}
\label{fig:appendix-il-real}
\end{figure}

\begin{figure}[!h]
\centering
\includegraphics[width=0.99\linewidth]{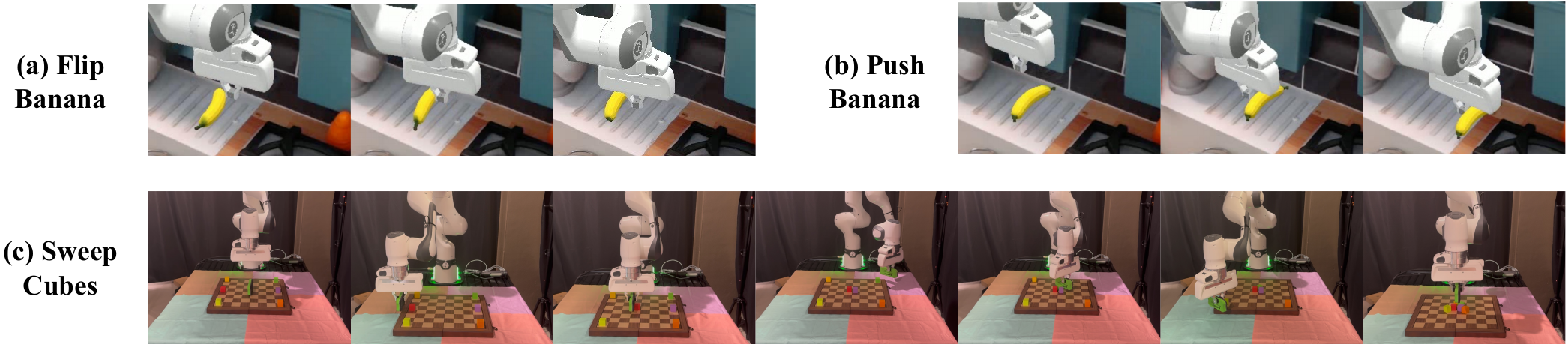}
\caption{Execution trajectories of dynamic manipulation tasks. (a) Flip Banana and (b) Push Banana are performed in simulation, demonstrating dynamic object interaction. (c) Sweep Cubes is executed in the real world and involves tool use, showcasing the pipeline’s ability to generalize to complex and contact-rich tasks.}
\label{fig:dynamic_exp}
\vspace{-8pt}
\end{figure}

Figure~\ref{fig:appendix-setup} illustrates the setup of our real-world experiment. Figure~\ref{fig:appendix-il-real} illustrates the comparison between RoLA-generated simulation data and the corresponding sim-to-real deployment in the real world on two tasks: \textit{Manipulation in Cluttered Scenes} and \textit{Pouring Water}. In addition to pick-and-place tasks, we include three dynamic tasks—one of which involves tool use—to demonstrate the versatility of our pipeline. As shown in Figure~\ref{fig:dynamic_exp}, we present execution trajectories for two simulated tasks: \textit{flip banana} and \textit{push banana}, and one real-world task: \textit{sweep cubes}.

\subsection{VLA Model Training \& Evaluation}

\begin{wrapfigure}{r}{0.5\textwidth}
\vspace{-10pt}
\centering
\includegraphics[width=1.\linewidth]{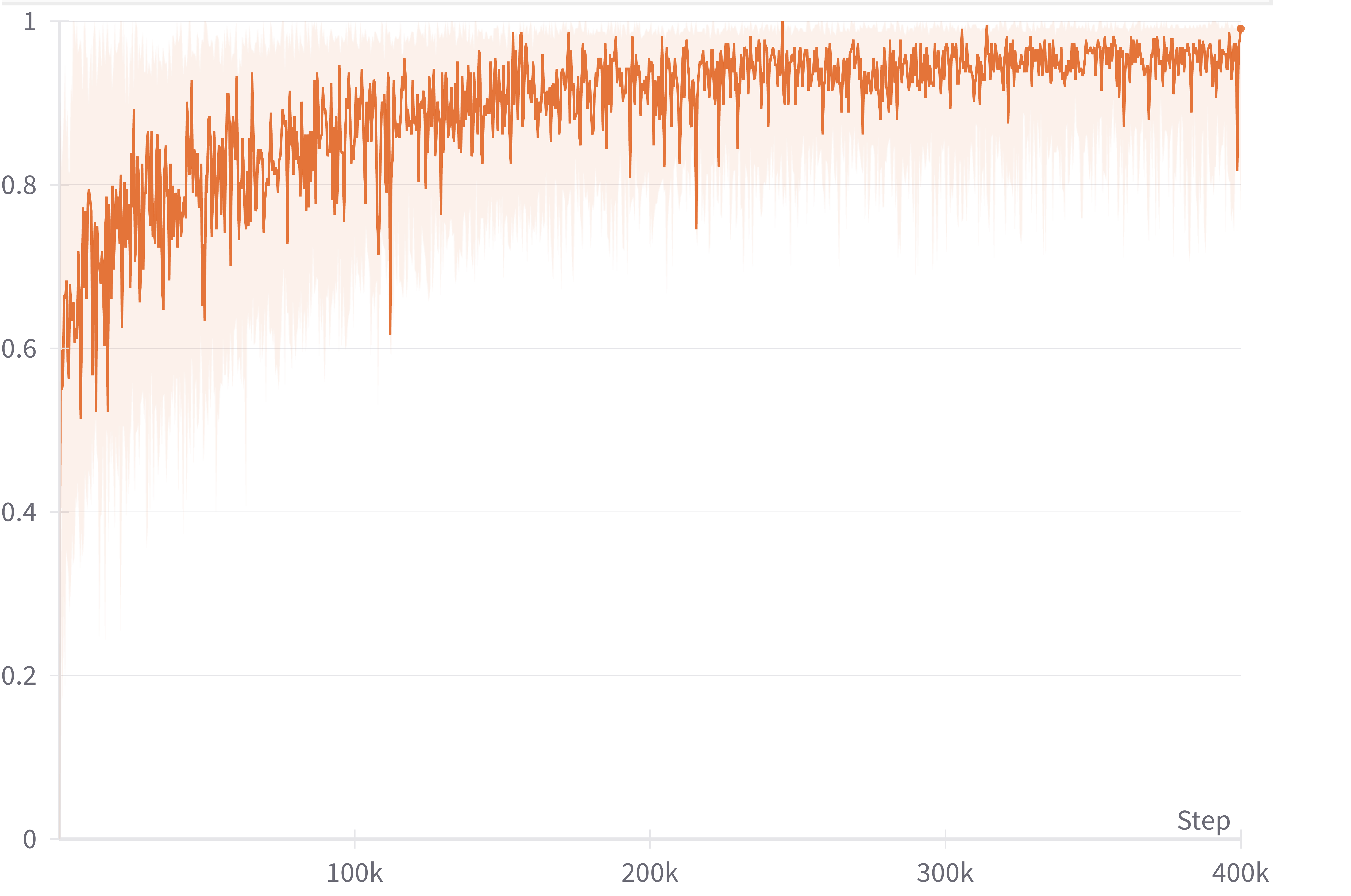}
\caption{\textbf{Training Curve of VLA.} Action token accuracy steadily increases and surpasses 95\% within 400K steps.}
\label{fig:appendix-vla-train}
\vspace{-14pt}
\end{wrapfigure}

\begin{figure}[!h]
\centering
\includegraphics[width=0.99\linewidth]{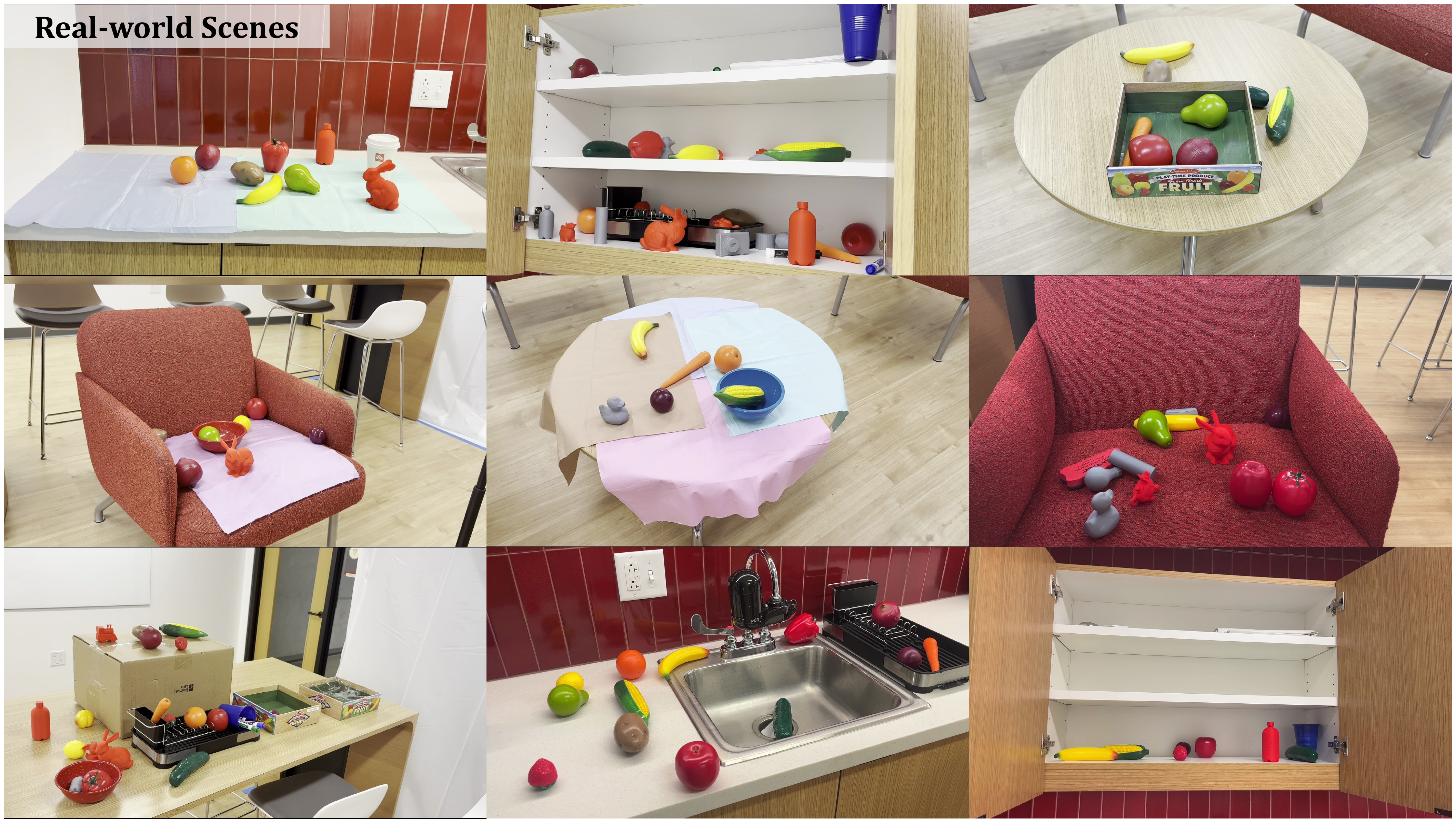}
\caption{\textbf{Example Real-World Scenes.} Sample images captured across diverse environments for vision-language task specification.}
\label{fig:appendix-vla-scene}
\end{figure}

To demonstrate the scalability of our data generation pipeline, we train a vision-language-action (VLA) model by first capturing real-world photographs and then transforming them into robotic training data. We target the same scale as BridgeData V2~\citep{bridgedata}. Specifically, we collect 2,000 real-world images across 12 distinct environments (compared to 24 in BridgeData V2), a process completed within 2 hours of human effort—the only human intervention required. Next, we use GPT-4o to propose approximately 10 tasks per image, yielding around 20,000 unique (image, language instruction) pairs. Our pipeline RoLA is then used to generate approximately 60,000 robotic trajectories, matching the scale of BridgeData V2. Figure~\ref{fig:appendix-vla-scene} displays example images captured in various environments.

For model architecture and training, we strictly follow the MiniVLA~\citep{minivla} setup. The model is trained for 400,000 steps until the action token accuracy exceeds 95\%. Training is performed on 8×H100 GPUs over four days (i.e., 768 GPU-hours). Figure~\ref{fig:appendix-vla-train} shows the accuracy progression over training steps.

\textbf{Evaluation.} For simulation-based evaluation, we closely follow the SimplerEnv~\citep{simplerenv} setup, with necessary modifications to the background image, manipulated object, camera pose, and language instruction to align with our task settings. We do not directly evaluate on the original SimplerEnv tasks because our model is not trained on any Open X-Embodiment~\citep{o2024open} real-world data, making such a comparison unfair. For the same reason, we do not compare our model against models like OpenVLA~\citep{openvla}, which are trained on Open X-Embodiment but not on any of our generated data—such a comparison would be unfair to OpenVLA.

We showcase two additional tasks in Table~\ref{tab:appendix_vla}, both using the same language instruction: \textit{pick up the banana}. These tasks are evaluated in distinct background environments to assess the model's ability to generalize. The results show that our model can robustly handle varying visual contexts while correctly executing the instructed behavior.

Below are descriptions of all tasks that requires \textit{partial success} (i.e., a score of 0.5):

\begin{enumerate}
\item \textbf{Pick up the Banana (Scene 2)}: Partial credit is given when the robot successfully \textit{grasps the banana} but fails to \textit{lift} it.
\item \textbf{Put the Green Pepper beside the Lemon}: Partial credit is given when the robot successfully \textit{picks up the green pepper} but does not perform the \textit{place} action.
\item \textbf{Take the Banana and Place it beside the Potato}: Partial credit is given when the robot successfully \textit{picks up the banana} but does not perform the \textit{place} action.
\item \textbf{Pick up the Red Tomato and Place it beside the Yellow Lemon}: Partial credit is given when the robot successfully \textit{picks up the red tomato} but does not perform the \textit{place} action.
\item \textbf{Take the Grey Object beside the Lemon and Place it beside the Yellow Apple}: Partial credit is given when the robot successfully \textit{picks up the grey object} but does not perform the \textit{place} action.
\end{enumerate}

\renewcommand{\arraystretch}{1.1}
\begin{table}[t!]
\vspace{-5mm}
\centering
\begin{tabular}{lc|lc}
\toprule
\textbf{Language Instruction} & \textbf{Success Rate} & \textbf{Language Instruction} & \textbf{Success Rate} \\
\hline
pick up the banana (scene 1) & 10/10 & pick up the banana (scene 2) & 7.5/10 \\
\bottomrule
\end{tabular}
\vspace{6pt}
\caption{Simulation evaluation of our VLA model trained on RoLA-generated data. Each task is tested over 10 trials with randomized object poses, under SimplerEnv~\citep{simplerenv} setting. Note that \textit{partial success} (score of 0.5) is possible for some tasks, following OpenVLA~\citep{openvla}.}
\label{tab:appendix_vla}
\vspace{-12pt}
\end{table}

\subsection{Learning from Internet Images}

\begin{figure}[!h]
\centering
\includegraphics[width=0.99\linewidth]{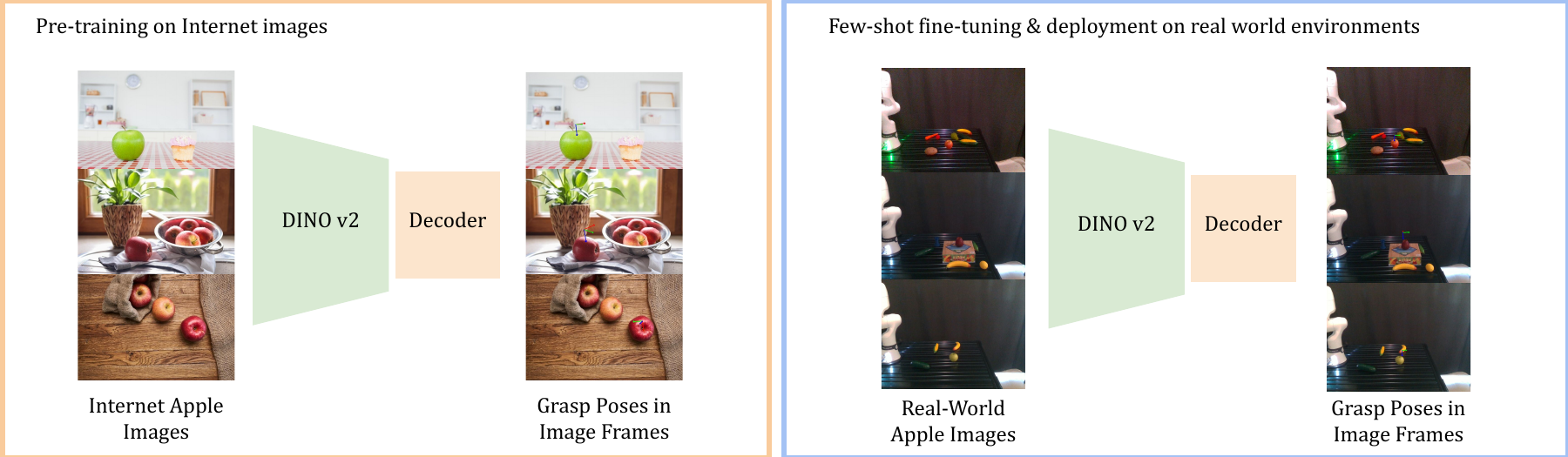}
\caption{The pretraining-finetuning paradigm for learning from Internet images.}
\label{fig:appendix-apple-prior}
\end{figure}

We investigate how Internet images can serve as a data source for robot learning, using apple picking as a case study due to the resource constraints. Our motivation is that Internet images exhibit diverse apple types, sizes, environments, and camera viewpoints. If we can generate grasping data from these images, the learned policy could generalize across object variations, different viewpoints, and scenarios. However, due to the huge domain gap between Internet images and real-world robotic environments, direct zero-shot deployment of Internet-learned policy to the real world is infeasible. Instead, we adopt a pretraining–finetuning paradigm: we first pretrain a grasping policy on Internet-generated apple grasping data, then finetune it using a small number of real robot demonstrations for alignment and fast policy adaptation. An illustration of the whole process is shown in Figure~\ref{fig:appendix-apple-prior}.

\textbf{Pretraining.} We collect massive Internet apple images, and leverage RoLA to generate robotic data for picking up the apples in images. We filter out those imperfect images and unsuccessful demonstrations, resulting in over 3000 high-quality demonstrations from Internet apple images. Since Internet images are captured from different camera viewpoints and the demonstrations are collected by different robot positions, to unify the data representation, we extract a data sample of an image and the corresponding 6-DoF apple grasp pose in the image frame for each demonstration, which is agnostic to robot positions. Then, we pretrain a neural network that takes an image as input and outputs the 6-DoF grasp pose in the image frame. The neural network contains a DINO v2 image encoder and a small head for decoding the apple grasp pose. The trained network will serve as an apple grasping prior for the subsequent real robot deployment.

\textbf{Finetuning.} In the real robotic environment, we collect 10/20/50 real robotic demonstrations for picking up an apple, extract the respective camera image and apple grasping pose in the image frame, and leverage these data to fine-tune the pretrained neural network. We set up a baseline by leveraging the same network architecture (DINO v2 + decoder) and directly training the network on the real robotic demonstrations without pretraining on the Internet apple images. During deployment, both the baseline and our network take in the image from the RealScene camera and output the apple grasping pose in the image frame. The grasping pose is then transformed into the robot base frame, and we leverage a motion planner to move the robot gripper to the estimated grasping pose to pick up the apple. We test the real-world grasping success rates and change the apple types, positions, and camera viewpoints during evaluation.

\begin{figure}[!h]
\centering
\includegraphics[width=0.95\linewidth]{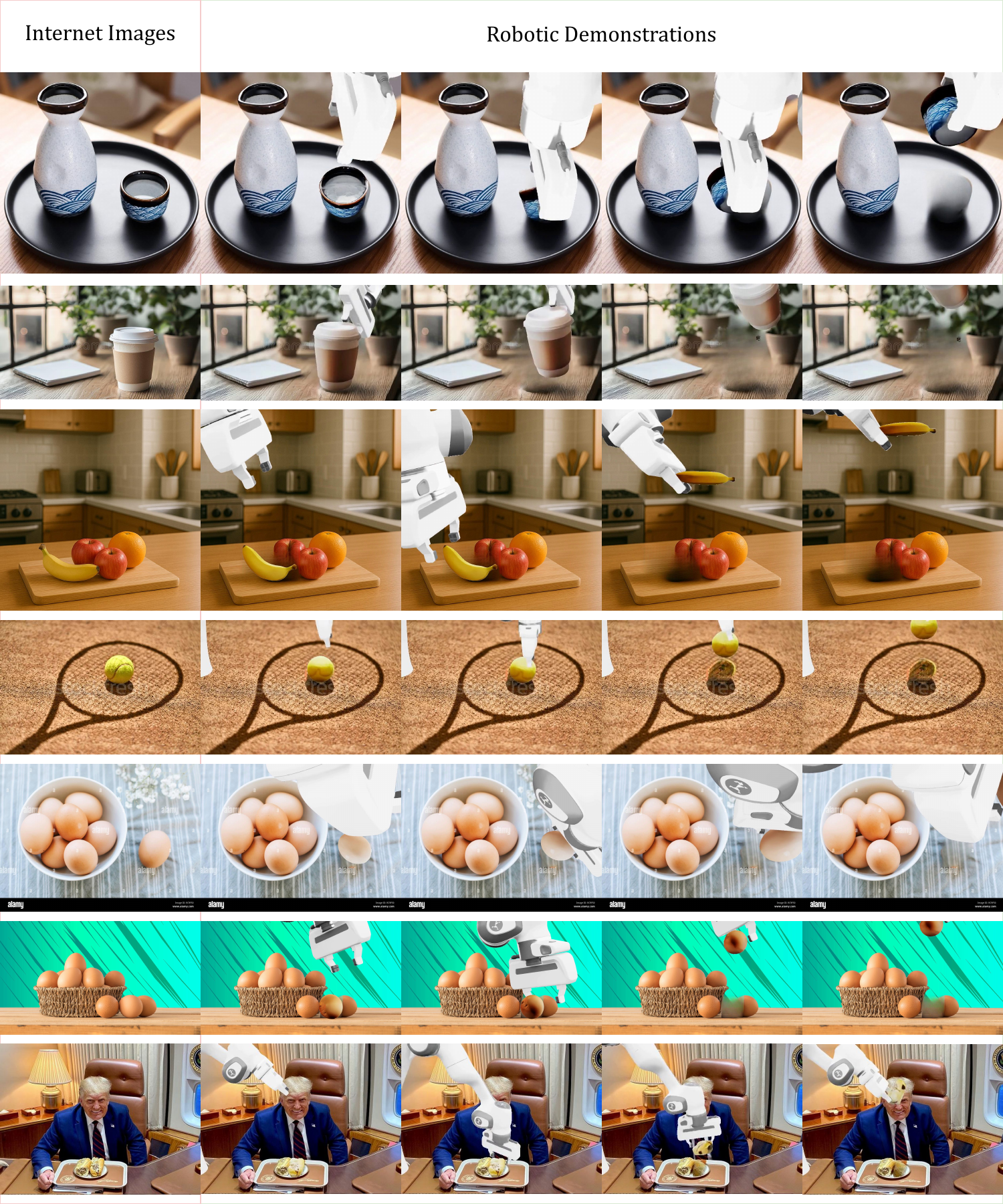}
\caption{Robotic demonstrations from Internet images.}
\label{fig:appendix-internet}
\end{figure}

\textbf{Robotic Demonstrations from Internet Images.} Figure~\ref{fig:appendix-internet} shows more examples of robotic demonstrations generated from other Internet images. Our approach is generalizable to different types of objects and images and enables robotic manipulation on in-the-wild images.

\section{Method}
\begin{figure}[!h]
    \centering
    \includegraphics[width=0.3\linewidth]{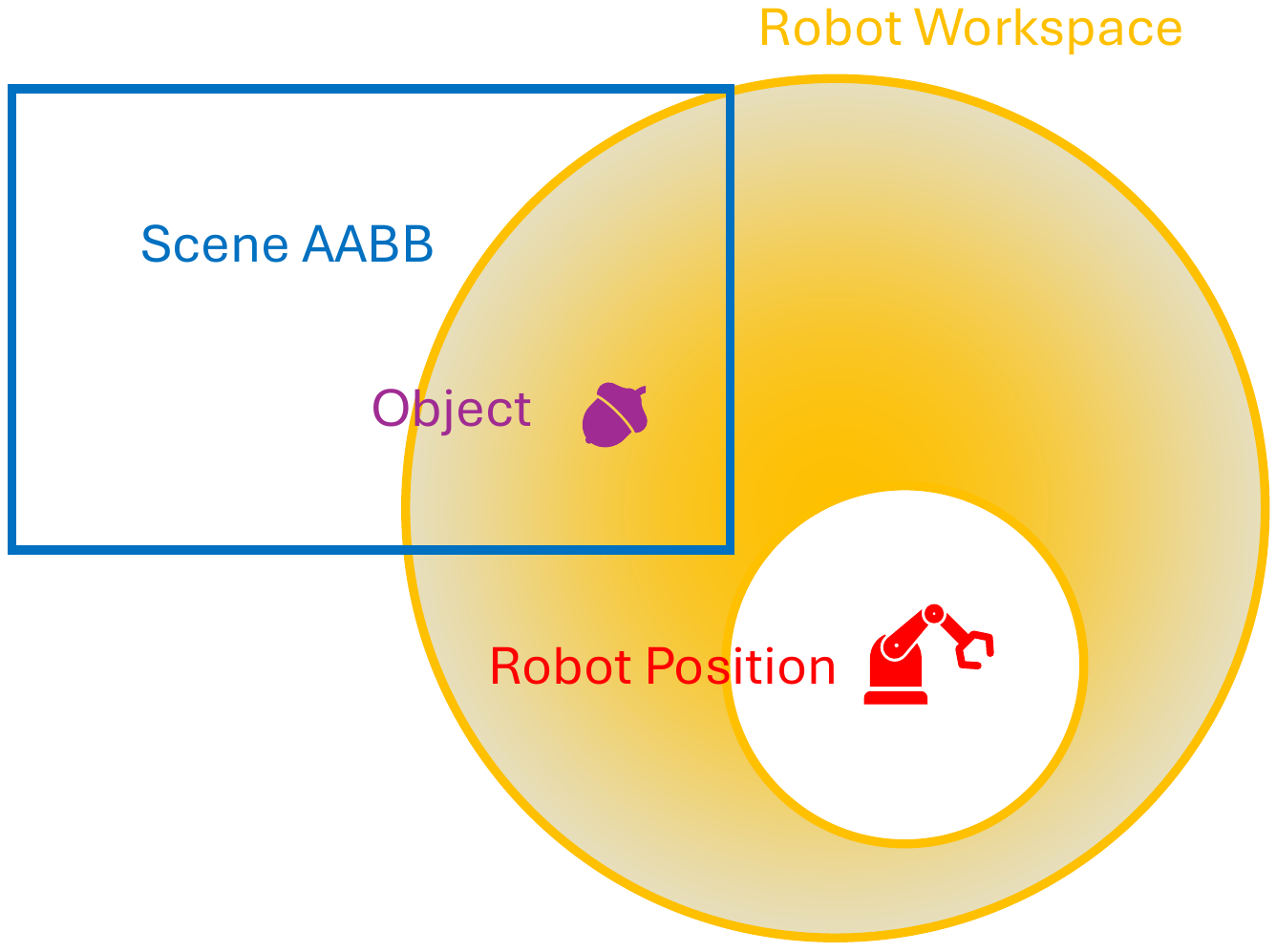}
    \caption{Visualization of the sampling-based method for generating feasible object placements.}
    \label{fig/robot_placement}
\end{figure}

\subsection{Background Geometry Modeling}

Generating background meshes from images is challenging, as removing objects will leave a hole in the background point clouds, and how to in-paint the geometry behind the object remains a problem. One solution is to have another depth estimation on the inpainted background image, but this introduces geometry errors due to the inconsistency of depth estimations. In this paper, we leverage the supported plane to complete the background geometry. Specifically, for pixels in the masked object regions, we shoot a ray from each pixel and check the intersection point of the ray and the supported plane. The intersection points will be used for creating background meshes. We also check the intersections of the ray and the scene's axis-aligned bounding boxes in case the ray does not intersect with the supported plane. 

\subsection{Robot Placement}
For non-robotic images, we propose a sampling-based method to generate feasible placements, as shown in Figure~\ref{fig/robot_placement}.

\subsection{Visual Blending}

\begin{figure}[!h]
\centering
\includegraphics[width=0.99\linewidth]{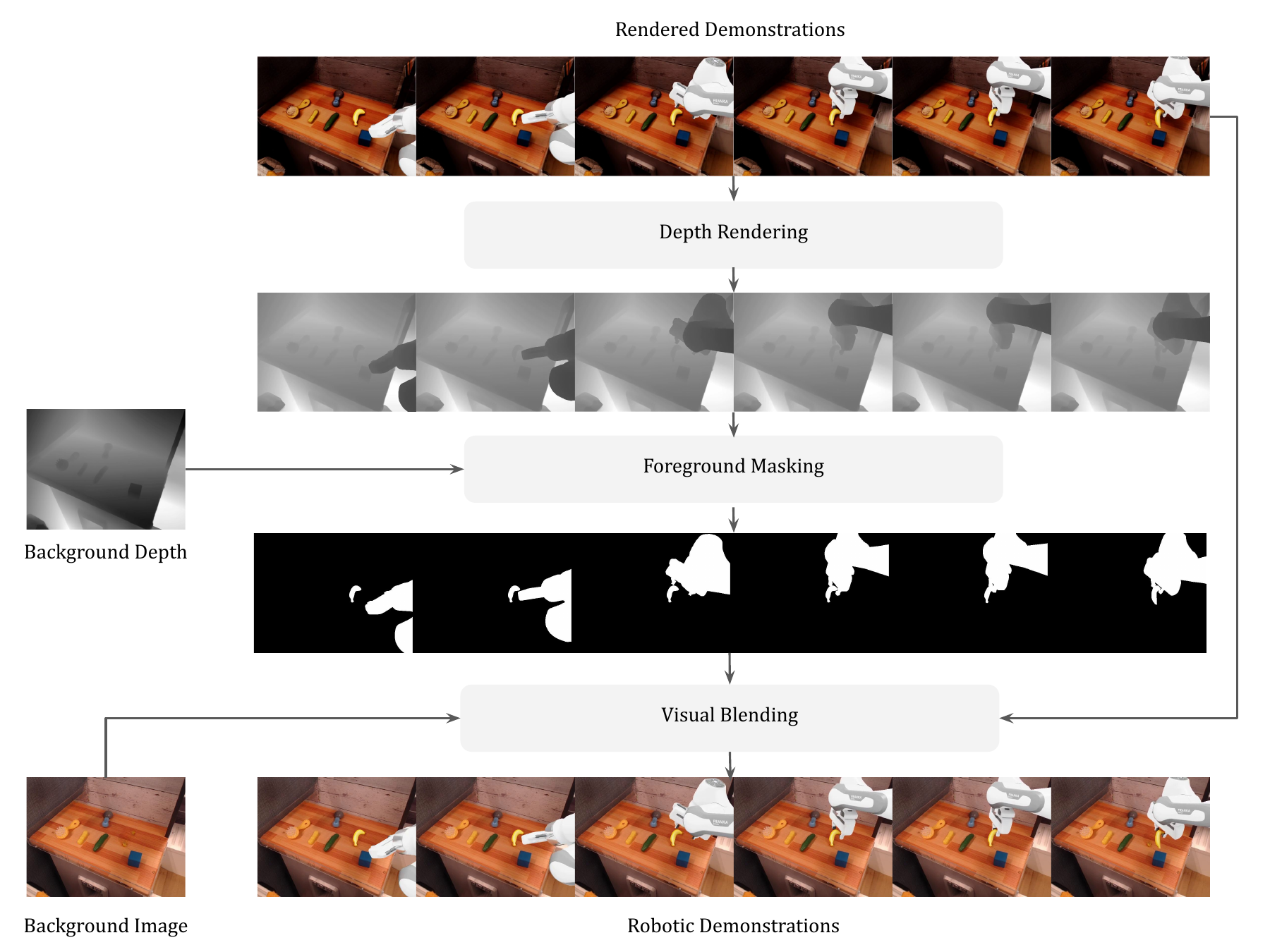}
\caption{An illustration of visual blending.}
\label{fig:appendix-blending}
\end{figure}

Figure~\ref{fig:appendix-blending} illustrates the detailed visual blending process. We generate foreground masks by comparing the rendered and background depths. For foreground regions, we set the pixels to those in the rendered frames. For background regions, we directly use the background image. This mitigates the rendering artifacts and leads to more photo-realistic results.

%% file: main.bbl
\begin{thebibliography}{64}
\providecommand{\natexlab}[1]{#1}
\providecommand{\url}[1]{\texttt{#1}}
\expandafter\ifx\csname urlstyle\endcsname\relax
  \providecommand{\doi}[1]{doi: #1}\else
  \providecommand{\doi}{doi: \begingroup \urlstyle{rm}\Url}\fi

\bibitem[Dai et~al.(2024)Dai, Wong, Jiang, Wang, Gokmen, Zhang, Wu, and Fei-Fei]{acdc}
T.~Dai, J.~Wong, Y.~Jiang, C.~Wang, C.~Gokmen, R.~Zhang, J.~Wu, and L.~Fei-Fei.
\newblock Automated creation of digital cousins for robust policy learning.
\newblock \emph{CoRL}, 2024.

\bibitem[Zook et~al.(2024)Zook, Sun, Spjut, Blukis, Birchfield, and Tremblay]{grs}
A.~Zook, F.-Y. Sun, J.~Spjut, V.~Blukis, S.~Birchfield, and J.~Tremblay.
\newblock Grs: Generating robotic simulation tasks from real-world images.
\newblock \emph{arXiv preprint arXiv:2410.15536}, 2024.

\bibitem[Han et~al.(2025)Han, Liu, Chen, Yu, Lyu, Tian, Wang, Zhang, and Pang]{re3sim}
X.~Han, M.~Liu, Y.~Chen, J.~Yu, X.~Lyu, Y.~Tian, B.~Wang, W.~Zhang, and J.~Pang.
\newblock Re3 sim: Generating high-fidelity simulation data via 3d-photorealistic real-to-sim for robotic manipulation.
\newblock \emph{arXiv preprint arXiv:2502.08645}, 2025.

\bibitem[Torne et~al.(2024)Torne, Simeonov, Li, Chan, Chen, Gupta, and Agrawal]{real2sim}
M.~Torne, A.~Simeonov, Z.~Li, A.~Chan, T.~Chen, A.~Gupta, and P.~Agrawal.
\newblock Reconciling reality through simulation: A real-to-sim-to-real approach for robust manipulation.
\newblock \emph{RSS}, 2024.

\bibitem[Fang et~al.(2025)Fang, Yang, Zhu, Zheng, Bertasius, Szafir, and Ding]{rebot}
Y.~Fang, Y.~Yang, X.~Zhu, K.~Zheng, G.~Bertasius, D.~Szafir, and M.~Ding.
\newblock Rebot: Scaling robot learning with real-to-sim-to-real robotic video synthesis.
\newblock \emph{arXiv preprint arXiv:2503.14526}, 2025.

\bibitem[Lou et~al.(2025)Lou, Liu, Pan, Geng, Chen, Ma, Li, Wang, Feng, Shi, et~al.]{robogs}
H.~Lou, Y.~Liu, Y.~Pan, Y.~Geng, J.~Chen, W.~Ma, C.~Li, L.~Wang, H.~Feng, L.~Shi, et~al.
\newblock Robo-gs: A physics consistent spatial-temporal model for robotic arm with hybrid representation.
\newblock \emph{ICRA}, 2025.

\bibitem[Li et~al.(2024)Li, Li, Zhang, Zhang, Jia, Wang, Fan, Tseng, and Wang]{robogsim}
X.~Li, J.~Li, Z.~Zhang, R.~Zhang, F.~Jia, T.~Wang, H.~Fan, K.-K. Tseng, and R.~Wang.
\newblock Robogsim: A real2sim2real robotic gaussian splatting simulator.
\newblock \emph{arXiv preprint arXiv:2411.11839}, 2024.

\bibitem[Pfaff et~al.(2025)Pfaff, Fu, Binagia, Isola, and Tedrake]{scalable-real2sim}
N.~Pfaff, E.~Fu, J.~Binagia, P.~Isola, and R.~Tedrake.
\newblock Scalable real2sim: Physics-aware asset generation via robotic pick-and-place setups.
\newblock \emph{arXiv preprint arXiv:2503.00370}, 2025.

\bibitem[Ye et~al.(2025)Ye, Liu, Ding, Gao, Rybkin, and Abbeel]{video2policy}
W.~Ye, F.~Liu, Z.~Ding, Y.~Gao, O.~Rybkin, and P.~Abbeel.
\newblock Video2policy: Scaling up manipulation tasks in simulation through internet videos.
\newblock \emph{arXiv preprint arXiv:2502.09886}, 2025.

\bibitem[Patel et~al.(2025)Patel, Yin, Huang, Garg, Nayyeri, Fei-Fei, Lazebnik, and Li]{vlm-real2sim}
S.~Patel, X.~Yin, W.~Huang, S.~Garg, H.~Nayyeri, L.~Fei-Fei, S.~Lazebnik, and Y.~Li.
\newblock A real-to-sim-to-real approach to robotic manipulation with vlm-generated iterative keypoint rewards.
\newblock \emph{ICRA}, 2025.

\bibitem[Zhu et~al.(2025)Zhu, Mou, Li, Ye, Huang, and Zhao]{vr-robo}
S.~Zhu, L.~Mou, D.~Li, B.~Ye, R.~Huang, and H.~Zhao.
\newblock Vr-robo: A real-to-sim-to-real framework for visual robot navigation and locomotion.
\newblock \emph{arXiv preprint arXiv:2502.01536}, 2025.

\bibitem[Chang et~al.(2017)Chang, Dai, Funkhouser, Halber, Niebner, Savva, Song, Zeng, and Zhang]{matterport3d}
A.~Chang, A.~Dai, T.~Funkhouser, M.~Halber, M.~Niebner, M.~Savva, S.~Song, A.~Zeng, and Y.~Zhang.
\newblock Matterport3d: Learning from rgb-d data in indoor environments.
\newblock In \emph{3DV}, 2017.

\bibitem[Xia et~al.(2018)Xia, Zamir, He, Sax, Malik, and Savarese]{gibson}
F.~Xia, A.~R. Zamir, Z.~He, A.~Sax, J.~Malik, and S.~Savarese.
\newblock Gibson env: Real-world perception for embodied agents.
\newblock In \emph{CVPR}, 2018.

\bibitem[Peng et~al.(2024)Peng, Lv, Zeng, Chen, Zhao, Sun, Lu, and Shao]{tiebot}
W.~Peng, J.~Lv, Y.~Zeng, H.~Chen, S.~Zhao, J.~Sun, C.~Lu, and L.~Shao.
\newblock Tiebot: Learning to knot a tie from visual demonstration through a real-to-sim-to-real approach.
\newblock \emph{CoRL}, 2024.

\bibitem[Chen et~al.(2024)Chen, Walsman, Memmel, Mo, Fang, Vemuri, Wu, Fox, and Gupta]{urdformer}
Z.~Chen, A.~Walsman, M.~Memmel, K.~Mo, A.~Fang, K.~Vemuri, A.~Wu, D.~Fox, and A.~Gupta.
\newblock Urdformer: A pipeline for constructing articulated simulation environments from real-world images.
\newblock \emph{RSS}, 2024.

\bibitem[Jiang et~al.(2022)Jiang, Hsu, and Zhu]{ditto}
Z.~Jiang, C.-C. Hsu, and Y.~Zhu.
\newblock Ditto: Building digital twins of articulated objects from interaction.
\newblock In \emph{CVPR}, 2022.

\bibitem[Nie et~al.(2023)Nie, Gadre, Ehsani, and Song]{structurefromaction}
N.~Nie, S.~Y. Gadre, K.~Ehsani, and S.~Song.
\newblock Structure from action: Learning interactions for articulated object 3d structure discovery.
\newblock \emph{IROS}, 2023.

\bibitem[Mandi et~al.(2025)Mandi, Weng, Bauer, and Song]{real2code}
Z.~Mandi, Y.~Weng, D.~Bauer, and S.~Song.
\newblock Real2code: Reconstruct articulated objects via code generation.
\newblock \emph{ICLR}, 2025.

\bibitem[Li et~al.(2023)Li, Zhang, Wong, Gokmen, Srivastava, Mart{\'\i}n-Mart{\'\i}n, Wang, Levine, Lingelbach, Sun, et~al.]{behavior1k}
C.~Li, R.~Zhang, J.~Wong, C.~Gokmen, S.~Srivastava, R.~Mart{\'\i}n-Mart{\'\i}n, C.~Wang, G.~Levine, M.~Lingelbach, J.~Sun, et~al.
\newblock Behavior-1k: A benchmark for embodied ai with 1,000 everyday activities and realistic simulation.
\newblock In \emph{CoRL}, 2023.

\bibitem[Kolve et~al.(2017)Kolve, Mottaghi, Han, VanderBilt, Weihs, Herrasti, Deitke, Ehsani, Gordon, Zhu, et~al.]{ai2thor}
E.~Kolve, R.~Mottaghi, W.~Han, E.~VanderBilt, L.~Weihs, A.~Herrasti, M.~Deitke, K.~Ehsani, D.~Gordon, Y.~Zhu, et~al.
\newblock Ai2-thor: An interactive 3d environment for visual ai.
\newblock \emph{arXiv preprint arXiv:1712.05474}, 2017.

\bibitem[Yang et~al.(2024)Yang, Sun, Weihs, VanderBilt, Herrasti, Han, Wu, Haber, Krishna, Liu, et~al.]{holodeck}
Y.~Yang, F.-Y. Sun, L.~Weihs, E.~VanderBilt, A.~Herrasti, W.~Han, J.~Wu, N.~Haber, R.~Krishna, L.~Liu, et~al.
\newblock Holodeck: Language guided generation of 3d embodied ai environments.
\newblock In \emph{CVPR}, 2024.

\bibitem[Wang et~al.(2024)Wang, Xian, Chen, Wang, Wang, Fragkiadaki, Erickson, Held, and Gan]{robogen}
Y.~Wang, Z.~Xian, F.~Chen, T.-H. Wang, Y.~Wang, K.~Fragkiadaki, Z.~Erickson, D.~Held, and C.~Gan.
\newblock Robogen: Towards unleashing infinite data for automated robot learning via generative simulation.
\newblock \emph{ICML}, 2024.

\bibitem[Nasiriany et~al.(2024)Nasiriany, Maddukuri, Zhang, Parikh, Lo, Joshi, Mandlekar, and Zhu]{robocasa}
S.~Nasiriany, A.~Maddukuri, L.~Zhang, A.~Parikh, A.~Lo, A.~Joshi, A.~Mandlekar, and Y.~Zhu.
\newblock Robocasa: Large-scale simulation of everyday tasks for generalist robots.
\newblock \emph{RSS}, 2024.

\bibitem[Singh et~al.(2024)Singh, Loquercio, Sferrazza, Wu, Qi, Abbeel, and Malik]{handobject}
H.~G. Singh, A.~Loquercio, C.~Sferrazza, J.~Wu, H.~Qi, P.~Abbeel, and J.~Malik.
\newblock Hand-object interaction pretraining from videos.
\newblock \emph{arXiv preprint arXiv:2409.08273}, 2024.

\bibitem[Chen et~al.(2024)Chen, Wang, Yang, and Liu]{object-dex}
Y.~Chen, C.~Wang, Y.~Yang, and C.~K. Liu.
\newblock Object-centric dexterous manipulation from human motion data.
\newblock \emph{CoRL}, 2024.

\bibitem[Bahl et~al.(2022)Bahl, Gupta, and Pathak]{human-to-robot}
S.~Bahl, A.~Gupta, and D.~Pathak.
\newblock Human-to-robot imitation in the wild.
\newblock \emph{CoRL}, 2022.

\bibitem[Mao et~al.(2024)Mao, Zhao, Song, Shi, Ye, Zhang, Geng, Malik, Guizilini, and Wang]{uh1}
J.~Mao, S.~Zhao, S.~Song, T.~Shi, J.~Ye, M.~Zhang, H.~Geng, J.~Malik, V.~Guizilini, and Y.~Wang.
\newblock Learning from massive human videos for universal humanoid pose control.
\newblock \emph{arXiv preprint arXiv:2412.14172}, 2024.

\bibitem[He et~al.(2024)He, Luo, He, Xiao, Zhang, Zhang, Kitani, Liu, and Shi]{omnih2o}
T.~He, Z.~Luo, X.~He, W.~Xiao, C.~Zhang, W.~Zhang, K.~Kitani, C.~Liu, and G.~Shi.
\newblock Omnih2o: Universal and dexterous human-to-humanoid whole-body teleoperation and learning.
\newblock \emph{CoRL}, 2024.

\bibitem[Yang et~al.(2024)Yang, Du, Ghasemipour, Tompson, Schuurmans, and Abbeel]{unisim}
M.~Yang, Y.~Du, K.~Ghasemipour, J.~Tompson, D.~Schuurmans, and P.~Abbeel.
\newblock Learning interactive real-world simulators.
\newblock \emph{ICLR}, 2024.

\bibitem[Ko et~al.(2024)Ko, Mao, Du, Sun, and Tenenbaum]{actionless-video}
P.-C. Ko, J.~Mao, Y.~Du, S.-H. Sun, and J.~B. Tenenbaum.
\newblock Learning to act from actionless videos through dense correspondences.
\newblock \emph{ICLR}, 2024.

\bibitem[Kuang et~al.(2024)Kuang, Ye, Geng, Mao, Deng, Guibas, Wang, and Wang]{ram}
Y.~Kuang, J.~Ye, H.~Geng, J.~Mao, C.~Deng, L.~Guibas, H.~Wang, and Y.~Wang.
\newblock Ram: Retrieval-based affordance transfer for generalizable zero-shot robotic manipulation.
\newblock \emph{CoRL}, 2024.

\bibitem[Chen et~al.(2025)Chen, Sun, Zhang, Pollefeys, and Leutenegger]{vidbot}
H.~Chen, B.~Sun, A.~Zhang, M.~Pollefeys, and S.~Leutenegger.
\newblock Vidbot: Learning generalizable 3d actions from in-the-wild 2d human videos for zero-shot robotic manipulation.
\newblock \emph{CVPR}, 2025.

\bibitem[Yuan et~al.(2025)Yuan, Joshi, Zhu, Su, Zhao, and Gao]{roboengine}
C.~Yuan, S.~Joshi, S.~Zhu, H.~Su, H.~Zhao, and Y.~Gao.
\newblock Roboengine: Plug-and-play robot data augmentation with semantic robot segmentation and background generation.
\newblock \emph{arXiv preprint arXiv:2503.18738}, 2025.

\bibitem[Chen et~al.(2024)Chen, Xu, Dharmarajan, Irshad, Cheng, Keutzer, Tomizuka, Vuong, and Goldberg]{rovi}
L.~Y. Chen, C.~Xu, K.~Dharmarajan, M.~Z. Irshad, R.~Cheng, K.~Keutzer, M.~Tomizuka, Q.~Vuong, and K.~Goldberg.
\newblock Rovi-aug: Robot and viewpoint augmentation for cross-embodiment robot learning.
\newblock \emph{CoRL}, 2024.

\bibitem[Lepert et~al.(2025)Lepert, Fang, and Bohg]{phantom}
M.~Lepert, J.~Fang, and J.~Bohg.
\newblock Phantom: Training robots without robots using only human videos.
\newblock \emph{arXiv preprint arXiv:2503.00779}, 2025.

\bibitem[Lepert et~al.(2024)Lepert, Doshi, and Bohg]{shadow}
M.~Lepert, R.~Doshi, and J.~Bohg.
\newblock Shadow: Leveraging segmentation masks for cross-embodiment policy transfer.
\newblock \emph{CoRL}, 2024.

\bibitem[Duan et~al.(2023)Duan, Wang, Shridhar, Fox, and Krishna]{ar2d2}
J.~Duan, Y.~R. Wang, M.~Shridhar, D.~Fox, and R.~Krishna.
\newblock Ar2-d2: Training a robot without a robot.
\newblock \emph{CoRL}, 2023.

\bibitem[Li et~al.(2024)Li, Hsu, Gu, Pertsch, Mees, Walke, Fu, Lunawat, Sieh, Kirmani, et~al.]{simplerenv}
X.~Li, K.~Hsu, J.~Gu, K.~Pertsch, O.~Mees, H.~R. Walke, C.~Fu, I.~Lunawat, I.~Sieh, S.~Kirmani, et~al.
\newblock Evaluating real-world robot manipulation policies in simulation.
\newblock \emph{CoRL}, 2024.

\bibitem[Teoh et~al.(2024)Teoh, Patidar, Ma, and James]{greenscreen}
E.~Teoh, S.~Patidar, X.~Ma, and S.~James.
\newblock Green screen augmentation enables scene generalisation in robotic manipulation.
\newblock \emph{arXiv preprint arXiv:2407.07868}, 2024.

\bibitem[Chen et~al.(2025)Chen, Jiang, Liu, Gupta, Li, Zhao, and Wang]{physgen3d}
B.~Chen, H.~Jiang, S.~Liu, S.~Gupta, Y.~Li, H.~Zhao, and S.~Wang.
\newblock Physgen3d: Crafting a miniature interactive world from a single image.
\newblock \emph{CVPR}, 2025.

\bibitem[Jiang et~al.(2025)Jiang, Hsu, Zhang, Yu, Wang, and Li]{phystwin}
H.~Jiang, H.-Y. Hsu, K.~Zhang, H.-N. Yu, S.~Wang, and Y.~Li.
\newblock Phystwin: Physics-informed reconstruction and simulation of deformable objects from videos.
\newblock \emph{arXiv preprint arXiv:2503.17973}, 2025.

\bibitem[Yang et~al.(2024)Yang, Jia, Zhi, and Huang]{physcene}
Y.~Yang, B.~Jia, P.~Zhi, and S.~Huang.
\newblock Physcene: Physically interactable 3d scene synthesis for embodied ai.
\newblock In \emph{CVPR}, 2024.

\bibitem[Chow et~al.(2025)Chow, Mao, Li, Seita, Guizilini, and Wang]{physbench}
W.~Chow, J.~Mao, B.~Li, D.~Seita, V.~Guizilini, and Y.~Wang.
\newblock Physbench: Benchmarking and enhancing vision-language models for physical world understanding.
\newblock \emph{ICLR}, 2025.

\bibitem[Liu et~al.(2024)Liu, Ren, Gupta, and Wang]{physgen}
S.~Liu, Z.~Ren, S.~Gupta, and S.~Wang.
\newblock Physgen: Rigid-body physics-grounded image-to-video generation.
\newblock In \emph{ECCV}, 2024.

\bibitem[Hsu et~al.(2025)Hsu, Lin, Zhai, Xia, and Wang]{autovfx}
H.-Y. Hsu, Z.-H. Lin, A.~Zhai, H.~Xia, and S.~Wang.
\newblock Autovfx: Physically realistic video editing from natural language instructions.
\newblock \emph{3DV}, 2025.

\bibitem[Yao et~al.(2025)Yao, Zhang, Yan, Zeng, Zhang, Xu, Yang, Gu, and Yu]{cast}
K.~Yao, L.~Zhang, X.~Yan, Y.~Zeng, Q.~Zhang, L.~Xu, W.~Yang, J.~Gu, and J.~Yu.
\newblock Cast: Component-aligned 3d scene reconstruction from an rgb image.
\newblock \emph{arXiv preprint arXiv:2502.12894}, 2025.

\bibitem[Ren et~al.(2024)Ren, Liu, Zeng, Lin, Li, Cao, Chen, Huang, Chen, Yan, et~al.]{groundedsam}
T.~Ren, S.~Liu, A.~Zeng, J.~Lin, K.~Li, H.~Cao, J.~Chen, X.~Huang, Y.~Chen, F.~Yan, et~al.
\newblock Grounded sam: Assembling open-world models for diverse visual tasks.
\newblock \emph{arXiv preprint arXiv:2401.14159}, 2024.

\bibitem[Long et~al.(2024)Long, Guo, Lin, Liu, Dou, Liu, Ma, Zhang, Habermann, Theobalt, et~al.]{wonder3d}
X.~Long, Y.-C. Guo, C.~Lin, Y.~Liu, Z.~Dou, L.~Liu, Y.~Ma, S.-H. Zhang, M.~Habermann, C.~Theobalt, et~al.
\newblock Wonder3d: Single image to 3d using cross-domain diffusion.
\newblock In \emph{CVPR}, 2024.

\bibitem[Suvorov et~al.(2022)Suvorov, Logacheva, Mashikhin, Remizova, Ashukha, Silvestrov, Kong, Goka, Park, and Lempitsky]{lama}
R.~Suvorov, E.~Logacheva, A.~Mashikhin, A.~Remizova, A.~Ashukha, A.~Silvestrov, N.~Kong, H.~Goka, K.~Park, and V.~Lempitsky.
\newblock Resolution-robust large mask inpainting with fourier convolutions.
\newblock In \emph{WACV}, 2022.

\bibitem[Bochkovskii et~al.(2025)Bochkovskii, Delaunoy, Germain, Santos, Zhou, Richter, and Koltun]{depthpro}
A.~Bochkovskii, A.~Delaunoy, H.~Germain, M.~Santos, Y.~Zhou, S.~R. Richter, and V.~Koltun.
\newblock Depth pro: Sharp monocular metric depth in less than a second.
\newblock \emph{ICLR}, 2025.

\bibitem[Besl and McKay(1992)]{icp}
P.~Besl and N.~D. McKay.
\newblock A method for registration of 3-d shapes.
\newblock \emph{IEEE T-PAMI}, 1992.

\bibitem[Hurst et~al.(2024)Hurst, Lerer, Goucher, Perelman, Ramesh, Clark, Ostrow, Welihinda, Hayes, Radford, et~al.]{gpt4o}
A.~Hurst, A.~Lerer, A.~P. Goucher, A.~Perelman, A.~Ramesh, A.~Clark, A.~Ostrow, A.~Welihinda, A.~Hayes, A.~Radford, et~al.
\newblock Gpt-4o system card.
\newblock \emph{arXiv preprint arXiv:2410.21276}, 2024.

\bibitem[Sundaralingam et~al.(2023)Sundaralingam, Hari, Fishman, Garrett, Van~Wyk, Blukis, Millane, Oleynikova, Handa, Ramos, et~al.]{curobo}
B.~Sundaralingam, S.~K.~S. Hari, A.~Fishman, C.~Garrett, K.~Van~Wyk, V.~Blukis, A.~Millane, H.~Oleynikova, A.~Handa, F.~Ramos, et~al.
\newblock curobo: Parallelized collision-free minimum-jerk robot motion generation.
\newblock \emph{arXiv preprint arXiv:2310.17274}, 2023.

\bibitem[Fang et~al.(2023)Fang, Wang, Fang, Gou, Liu, Yan, Liu, Xie, and Lu]{anygrasp}
H.-S. Fang, C.~Wang, H.~Fang, M.~Gou, J.~Liu, H.~Yan, W.~Liu, Y.~Xie, and C.~Lu.
\newblock Anygrasp: Robust and efficient grasp perception in spatial and temporal domains.
\newblock \emph{IEEE T-RO}, 2023.

\bibitem[Fang et~al.(2020)Fang, Wang, Gou, and Lu]{graspnet}
H.-S. Fang, C.~Wang, M.~Gou, and C.~Lu.
\newblock Graspnet-1billion: A large-scale benchmark for general object grasping.
\newblock In \emph{CVPR}, 2020.

\bibitem[Walke et~al.(2023)Walke, Black, Zhao, Vuong, Zheng, Hansen-Estruch, He, Myers, Kim, Du, et~al.]{bridgedata}
H.~R. Walke, K.~Black, T.~Z. Zhao, Q.~Vuong, C.~Zheng, P.~Hansen-Estruch, A.~W. He, V.~Myers, M.~J. Kim, M.~Du, et~al.
\newblock Bridgedata v2: A dataset for robot learning at scale.
\newblock In \emph{CoRL}, 2023.

\bibitem[Chi et~al.(2023)Chi, Xu, Feng, Cousineau, Du, Burchfiel, Tedrake, and Song]{diffusionppolicy}
C.~Chi, Z.~Xu, S.~Feng, E.~Cousineau, Y.~Du, B.~Burchfiel, R.~Tedrake, and S.~Song.
\newblock Diffusion policy: Visuomotor policy learning via action diffusion.
\newblock \emph{IJRR}, 2023.

\bibitem[Kim et~al.(2024)Kim, Pertsch, Karamcheti, Xiao, Balakrishna, Nair, Rafailov, Foster, Lam, Sanketi, Vuong, Kollar, Burchfiel, Tedrake, Sadigh, Levine, Liang, and Finn]{openvla}
M.~J. Kim, K.~Pertsch, S.~Karamcheti, T.~Xiao, A.~Balakrishna, S.~Nair, R.~Rafailov, E.~Foster, G.~Lam, P.~Sanketi, Q.~Vuong, T.~Kollar, B.~Burchfiel, R.~Tedrake, D.~Sadigh, S.~Levine, P.~Liang, and C.~Finn.
\newblock Openvla: An open-source vision-language-action model, 2024.
\newblock URL \url{https://arxiv.org/abs/2406.09246}.

\bibitem[Yang et~al.(2024)Yang, Yang, Zhang, Hui, Zheng, Yu, Li, Liu, Huang, Wei, et~al.]{qwen}
A.~Yang, B.~Yang, B.~Zhang, B.~Hui, B.~Zheng, B.~Yu, C.~Li, D.~Liu, F.~Huang, H.~Wei, et~al.
\newblock Qwen2. 5 technical report.
\newblock \emph{arXiv preprint arXiv:2412.15115}, 2024.

\bibitem[Belkhale and Sadigh(2024)]{minivla}
S.~Belkhale and D.~Sadigh.
\newblock Minivla: A better vla with a smaller footprint.
\newblock 2024.

\bibitem[Ren et~al.(2024)Ren, Liu, Zeng, Lin, Li, Cao, Chen, Huang, Chen, Yan, Zeng, Zhang, Li, Yang, Li, Jiang, and Zhang]{ren2024grounded}
T.~Ren, S.~Liu, A.~Zeng, J.~Lin, K.~Li, H.~Cao, J.~Chen, X.~Huang, Y.~Chen, F.~Yan, Z.~Zeng, H.~Zhang, F.~Li, J.~Yang, H.~Li, Q.~Jiang, and L.~Zhang.
\newblock Grounded sam: Assembling open-world models for diverse visual tasks, 2024.

\bibitem[Tschannen et~al.(2025)Tschannen, Gritsenko, Wang, Naeem, Alabdulmohsin, Parthasarathy, Evans, Beyer, Xia, Mustafa, et~al.]{tschannen2025siglip}
M.~Tschannen, A.~Gritsenko, X.~Wang, M.~F. Naeem, I.~Alabdulmohsin, N.~Parthasarathy, T.~Evans, L.~Beyer, Y.~Xia, B.~Mustafa, et~al.
\newblock Siglip 2: Multilingual vision-language encoders with improved semantic understanding, localization, and dense features.
\newblock \emph{arXiv preprint arXiv:2502.14786}, 2025.

\bibitem[Wu et~al.(2023)Wu, Zhang, Fu, Wang, Jiawei~Ren, Wu, Yang, Wang, Qian, Lin, and Liu]{wu2023omniobject3d}
T.~Wu, J.~Zhang, X.~Fu, Y.~Wang, L.~P. Jiawei~Ren, W.~Wu, L.~Yang, J.~Wang, C.~Qian, D.~Lin, and Z.~Liu.
\newblock Omniobject3d: Large-vocabulary 3d object dataset for realistic perception, reconstruction and generation.
\newblock In \emph{IEEE/CVF Conference on Computer Vision and Pattern Recognition (CVPR)}, 2023.

\bibitem[O’Neill et~al.(2024)O’Neill, Rehman, Maddukuri, Gupta, Padalkar, Lee, Pooley, Gupta, Mandlekar, Jain, et~al.]{o2024open}
A.~O’Neill, A.~Rehman, A.~Maddukuri, A.~Gupta, A.~Padalkar, A.~Lee, A.~Pooley, A.~Gupta, A.~Mandlekar, A.~Jain, et~al.
\newblock Open x-embodiment: Robotic learning datasets and rt-x models: Open x-embodiment collaboration 0.
\newblock In \emph{2024 IEEE International Conference on Robotics and Automation (ICRA)}, pages 6892--6903. IEEE, 2024.

\end{thebibliography}
